\documentclass{article}

\PassOptionsToPackage{numbers, compress}{natbib}
\usepackage[final]{neurips_2022}
\usepackage{natbib}
\setcitestyle{nonatbib}

\usepackage[utf8]{inputenc} 
\usepackage[T1]{fontenc}    
\usepackage{hyperref}       
\usepackage{url}            
\usepackage{booktabs}       
\usepackage{amsfonts}       
\usepackage{nicefrac}       
\usepackage{microtype}      
\usepackage{xcolor}         

\usepackage{epsfig}
\usepackage{graphicx}
\usepackage{amsmath}
\usepackage{amssymb}
\usepackage{multirow}
\usepackage{caption}

\newcommand{\tabincell}[2]{\begin{tabular}{@{}#1@{}}#2\end{tabular}} 

\usepackage{mathrsfs}
\usepackage{subfigure}

\usepackage{paralist} 

\makeatletter 
\newcommand\figcaption{\def\@captype{figure}\caption} 
\newcommand\tabcaption{\def\@captype{table}\caption} 
\makeatother

\title{Margin-Based Few-Shot Class-Incremental Learning with Class-Level Overfitting Mitigation}

\author{%
  Yixiong Zou$^1$, Shanghang Zhang$^2$, Yuhua Li$^1$ and Ruixuan Li$^1$\thanks{Corresponding author.} \\
  $^1$School of Computer Science and Technology, Huazhong University of Science and Technology\\
  $^2$School of Computer Science, Peking University\\
  \texttt{$^1$\{yixiongz, idcliyuhua, rxli\}@hust.edu.cn, $^2$shanghang@pku.edu.cn} \\
}

\begin{document}

\maketitle

\begin{abstract}
  Few-shot class-incremental learning (FSCIL) is designed to incrementally recognize novel classes with only few training samples after the (pre-)training on base classes with sufficient samples, which focuses on both base-class performance and novel-class generalization. 
  A well known modification to the base-class training is to apply a margin to the base-class classification. 
  However, a dilemma exists that we can hardly achieve both good base-class performance and novel-class generalization simultaneously by applying the margin during the base-class training, which is still under explored. 
  In this paper, we study the cause of such dilemma for FSCIL. We first interpret this dilemma as a class-level overfitting (CO) problem from the aspect of pattern learning, and then find its cause lies in the easily-satisfied constraint of learning margin-based patterns. Based on the analysis, we propose a novel margin-based FSCIL method to mitigate the CO problem by providing the pattern learning process with extra constraint from the margin-based patterns themselves. Extensive experiments on CIFAR100, Caltech-USCD Birds-200-2011 (CUB200), and \textit{mini}ImageNet demonstrate that the proposed method effectively mitigates the CO problem and achieves state-of-the-art performance.
\end{abstract}



\section{Introduction}
\label{sec:intro}

With the development of deep learning, deep neural networks gradually demonstrate superior performance on the recognition of pre-defined classes with large amount of training data~\cite{simonyan2014very,he2016deep}. 
However, the model's generalization capability on the downstream novel classes is much less explored and still needs to be improved~\cite{kornblith2021better,guobroader}.
To deal with this problem, the few-shot class-incremental learning (FSCIL) task~\cite{hou2019learning,rebuffi2017icarl,castro2018end,tao2020few,zhang2021few,zhou2022forward} comes into sight. FSCIL first (pre-)trains a model on a set of pre-defined classes (base classes), and then generalizes the model to the incremental novel classes with only few training samples, simulating human's ability of continually learning novel concepts with only few examples, and emphasizing both the performance on the pre-defined base classes and the generalization on the downstream novel classes.

\begin{figure}[!t]
	\begin{tabular*}{\textwidth}{ccccc}
		& No Margin & Positive Margin & Negative Margin & Ours \\[0.05cm]
		\rotatebox{90}{Base Class}
		& \fbox{\includegraphics[width=0.2\linewidth,height=0.12\linewidth]{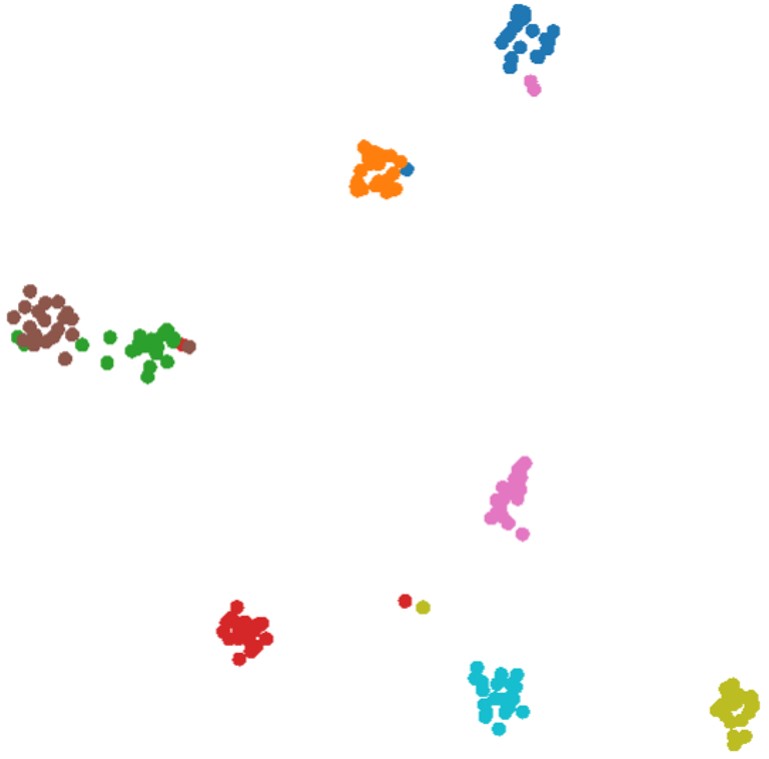}}\hspace{-0.15cm}
		& \fbox{\includegraphics[width=0.2\linewidth,height=0.12\linewidth]{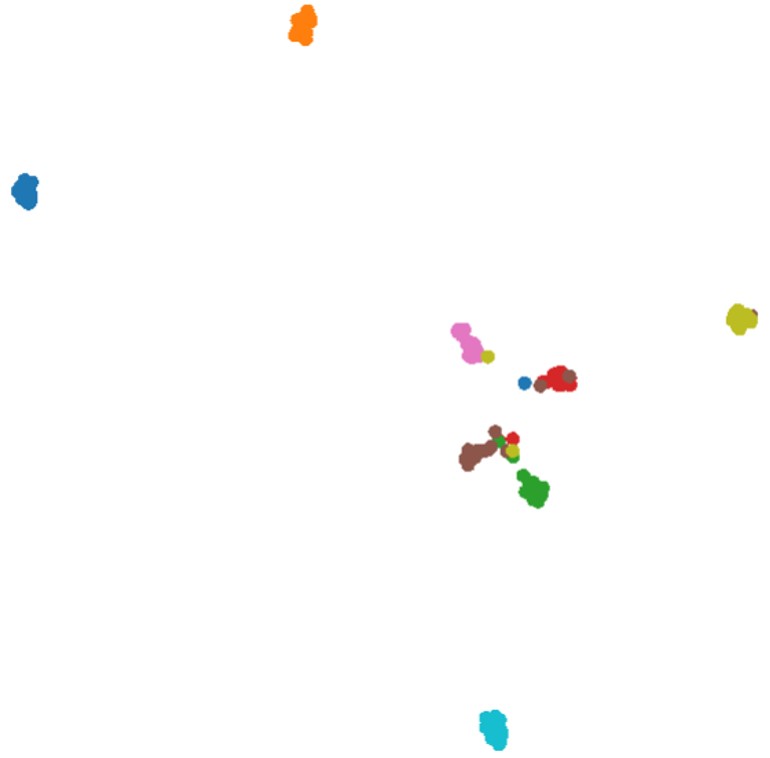}}\hspace{-0.15cm}
		& \fbox{\includegraphics[width=0.2\linewidth,height=0.12\linewidth]{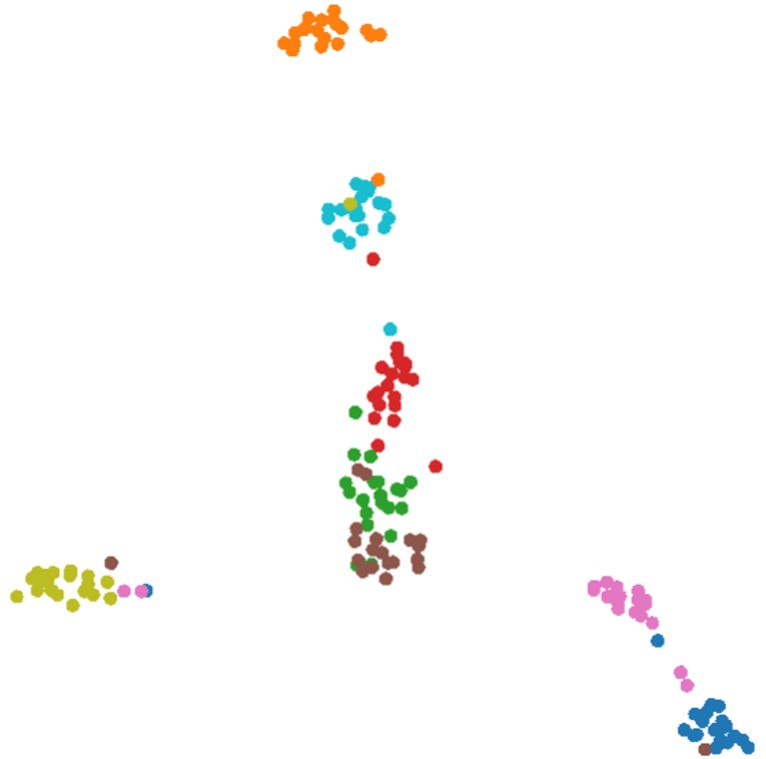}}\hspace{-0.15cm}
		& \fbox{\includegraphics[width=0.2\linewidth,height=0.12\linewidth]{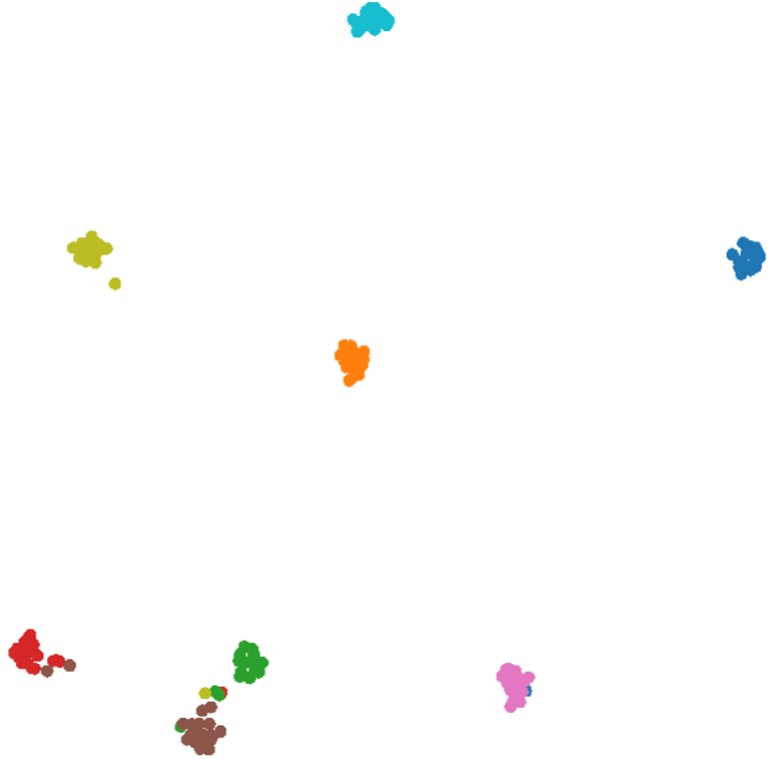}}
		\\[0.1cm]
		\rotatebox{90}{Novel Class} 
		& \fbox{\includegraphics[width=0.2\linewidth,height=0.12\linewidth]{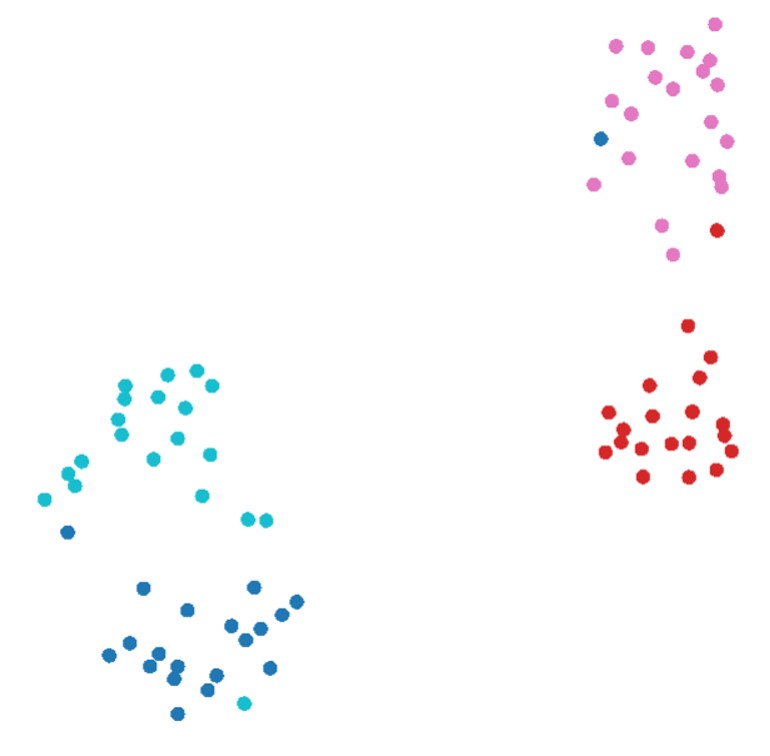}}\hspace{-0.15cm}
		& \fbox{\includegraphics[width=0.2\linewidth,height=0.12\linewidth]{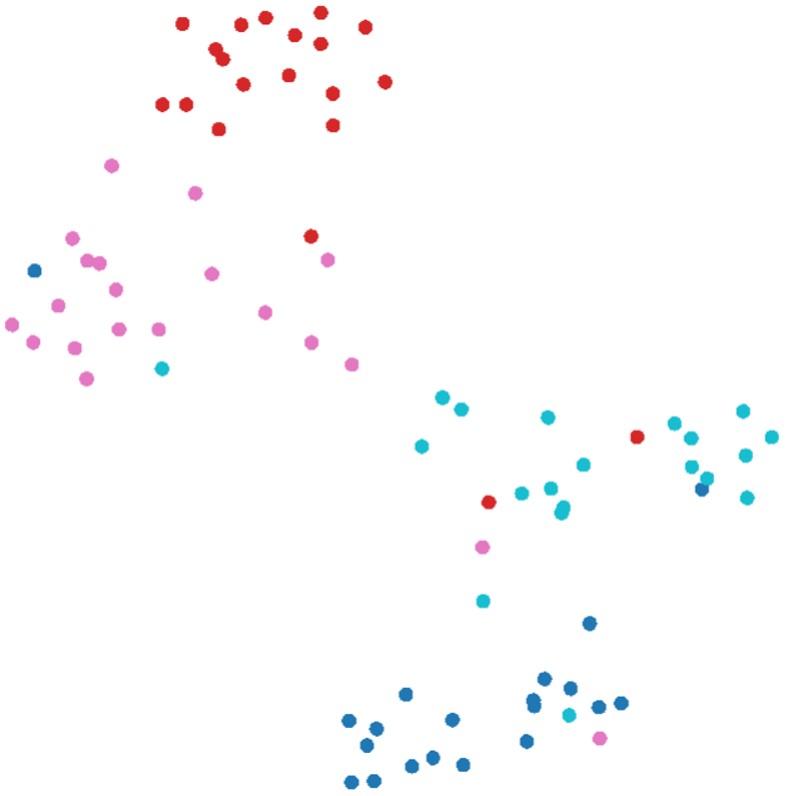}}\hspace{-0.15cm}
		& \fbox{\includegraphics[width=0.2\linewidth,height=0.12\linewidth]{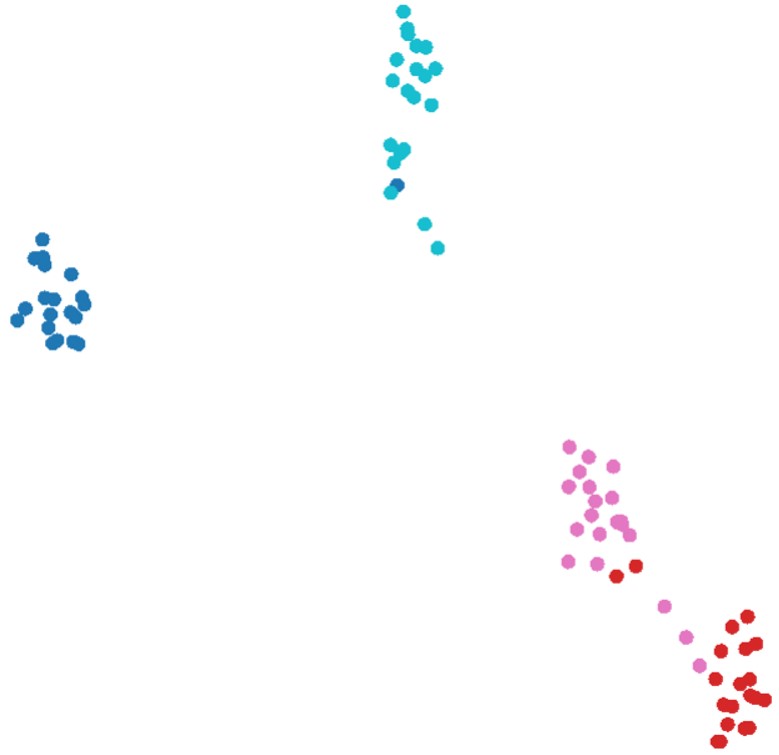}}\hspace{-0.15cm}
		& \fbox{\includegraphics[width=0.2\linewidth,height=0.12\linewidth]{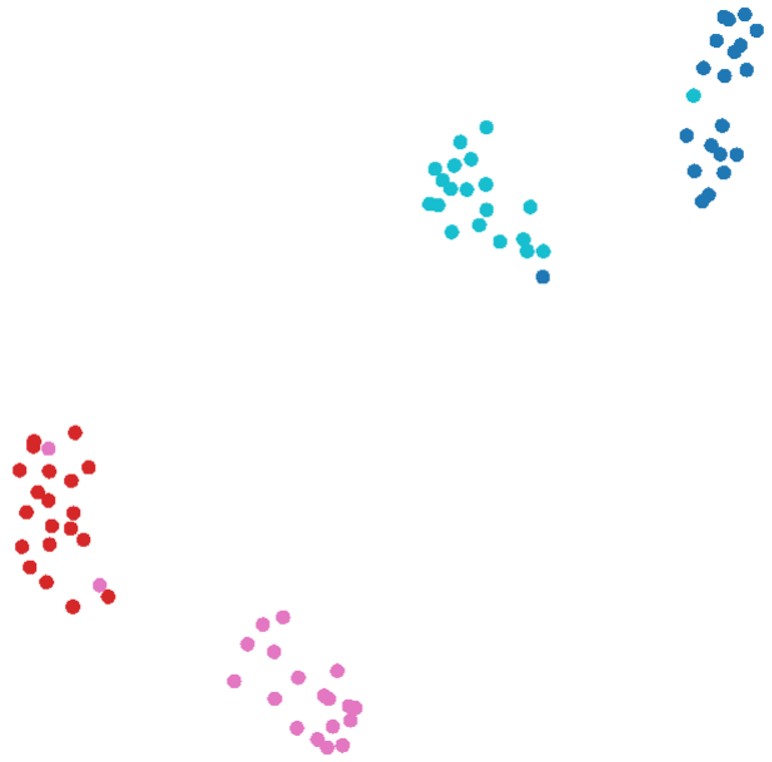}}
	\end{tabular*} \vspace{-0.2cm}
	\caption{A dilemma exists between base-class performance and downstream novel-class generalization. By applying positive margins, base-class features are better separated which indicates better base-class performance, but the novel-class features are confused which indicates lower novel-class generalization. In contrast, by applying negative margins, base-class features are confused but the novel-class features are better separated. In this paper, we study the cause of such dilemma for the few-shot class-incremental learning task, and propose a method to mitigate such dilemma to better separate both base and novel classes.}\vspace{-0.4cm}
	\label{fig:motivation} 
\end{figure}

However, a dilemma is recently revealed~\cite{kornblith2021better,chen2019transferability,cui2022discriminability,cui2020towards} that better loss functions, which lead to higher performance on the pre-training data, could lead to worse generalization on the downstream tasks.
As introduced by \cite{liu2020negative} and depicted in Fig.~\ref{fig:motivation}, similar phenomenon also exists in the FSCIL task that a positive classification margin~\cite{liu2020negative,wang2018cosface,deng2019arcface,sun2020circle} applied to the classification of the base-class (pre-)training could lead to higher base-class performance but lower novel-class performance, while a negative margin could result in lower base-class performance but increase the novel-class performance.
Although this dilemma widely exists in the tasks involving novel-class generalization such as few-shot learning (FSL) and FSCIL, only few works~\cite{liu2020negative} tried to explore its cause, and can hardly be used to handle it. Due to space limitation, we will provide extended related works in the appendix.

%
%
%

In this paper, we study the cause of the dilemma in the margin-based classification for the FSCIL problem from the the aspect of pattern learning. We find this dilemma can be understood as a class-level overfitting (CO) problem, which can be interpreted by the fitness of the learned patterns to each base class. The fitness determines how much the learned patterns are specific to some base classes or shared among classes, making the learned patterns either discriminative (tend to overfit base classes) or transferable (tend to underfit base classes) and causes the dilemma. Based on the interpretation, we discover the cause of the dilemma lies in the easily-satisfied constraint of learning shared or class-specific patterns. Therefore, we further design a novel FSCIL method to mitigate the dilemma of CO by providing the pattern learning process with extra constraint from margin-based patterns themselves, improving performance on both base and novel classes as shown in Fig.~\ref{fig:motivation}, and achieving state-of-the-art performance in terms of the all-class accuracy. Our contributions are:

\vspace{-0.2cm}
\begin{itemize}[$\bullet$]
	\item We interpret the dilemma of the margin-based classification as a class-level overfitting problem from the aspect of pattern learning.
	
	\item We find the cause of the class-level overfitting problem lies in the easily-satisfied constraint of learning shared or class-specific patterns.
	
	\item We propose a novel FSCIL method to mitigate the class-level overfitting problem based on the interpretation and analysis of the cause.
	
	\item Extensive experiments on three public datasets verify the rationale of the model design, and show that we can achieve state-of-the-art performance.
\end{itemize}

\vspace{-0.3cm}
\section{Interpreting the Dilemma of Few-Shot Class-Incremental Learning}
\label{sec:task_and_baseline}
\vspace{-0.2cm}

In this section, we first describe the Few-Shot Class-Incremental Learning (FSCIL) task and the baseline model, and then conduct experiments to analyze the dilemma.

\vspace{-0.2cm}
\subsection{Task and Baseline Description}
\vspace{-0.2cm}

The FSCIL task aims to incrementally recognize novel classes with only few training samples. 
Basically, the model is first (pre-)trained on a set of base classes with sufficient training samples (a.k.a. base session), then confronted with novel classes with limited training samples (a.k.a. incremental session), and finally required to recognize test samples from all encountered classes. 

Specifically, given the base session dataset $D^0=\{(x_i, y_i)\}_{i=1}^{n_0}$ with the label space $Y_0$, the model is trained to recognize all $|Y_0|$ classes from $Y_0$ by minimizing the loss
\vspace{-0.1cm}
\begin{equation}
	\sum_{(x_i, y_i) \in D^0} L(\phi(x_i), y_i),
	\label{eq:cls}
\end{equation}
where $L(\cdot,\cdot)$ is typically a cross-entropy loss, $\phi(\cdot)$ is the predictor which is composed of a backbone network $f(\cdot)$ for feature extraction and a linear classifier, represented as $\phi(x) = W^\top f(x)$ where $\phi(x) \in R^{N_0 \times 1}$, $W \in R^{d \times N_0}$ and $f(x) \in R^{d \times 1}$. Typically, $f(x)$ and the $W$ are $L_2$ normalized~\cite{zhang2021few}.

When the $k$th incremental session comes, the model needs to learn from its training data $D^k=\{(x_i, y_i)\}_{i=1}^{n_k}$. The weight of the classifier will be extended to represent the novel label space $Y_k$ imported by this session, represented as $W = \{w^0_1, w^0_2, ..., w^0_{|Y_0|}\} \cup ... \cup \{w^k_1, ..., w^k_{|Y_k|}\}$ where $w^k_j$ denotes the weight of the classifier corresponding to the $j$th class of the $k$th session. 

A strong baseline~\cite{zhang2021few} is to freeze model's parameters to avoid the catastrophic forgetting brought by the finetuning on novel-classes. For the incremental sessions (i.e., $k$ > 0), the average of the features extracted from the training data will be used as the classifier's weight~\cite{zhang2021few} (a.k.a. prototype) as $w^k_j = \frac{1}{n_k^j} \sum_{i=1}^{n_k^j} f(x_i)$, where $n_k^j$ denotes the number of training samples in the class $j$ for the session $k$. As this baseline focuses on the base-class training, in this paper, the term \textit{training}, if not otherwise stated, refers to the base-class training.
Finally, the performance of the $k$th session will be obtained by classifying the test samples from all $\sum_{i=0}^{k} |Y_i|$ encountered classes.


\vspace{-0.3cm}
\subsection{Margin-Based Classification}
\vspace{-0.2cm}
A well known modification to base-class training loss (Eq.~\ref{eq:cls}) is to integrate a margin~\cite{liu2020negative,deng2019arcface,wang2018cosface} as
\vspace{-0.1cm}
\begin{equation}
	L(x_i, y_i) = -log \frac{e^{\tau (w_{y_i} f(x_i) - m)}}{e^{\tau (w_{y_i} f(x_i) - m)} + \sum_{j \neq y_i} e^{\tau w_j f(x_i)}},
	\label{eq:margin}
\end{equation}
where $w_{y_i}$ refers to the classifier weight for class $y_i$, $\tau$ is typically set to 16.0 and $m$ is the margin.

As analyzed in \cite{liu2020negative}, empirically a dilemma exists that a positive margin could improve the base-class performance but harm the novel-class generalization, and reversely, a negative margin could contribute to the novel-class performance but decrease that of the base classes as shown in Fig.~\ref{fig:motivation}. Similar phenomenon has been observed in other works such as \cite{kornblith2021better} that a better loss function for the pre-training task could harm the generalization on downstream tasks.

\vspace{-0.3cm}
\subsection{Interpretation of Class-Level Overfitting from Pattern Learning View}
\vspace{-0.2cm}

Experiments are conducted on the CIFAR100~\cite{krizhevsky2009learning} dataset and reported in Fig.~\ref{fig:analysis_of_baseline}.
CIFAR100 contains 100 classes in all. As split by \cite{tao2020few}, 60 classes are chosen as base classes, and the remaining 40 classes (with 5 training samples in each class) are chosen as novel classes\footnote{\label{foot:appendix}Please refer to the appendix for details.}. Experiments are conducted on the last incremental session, where all 100 classes are involved.

From Fig.~\ref{fig:analysis_of_baseline} (left), we can see that as the margin increases, the base-class accuracy increases while the novel-class accuracy decreases, which is consistent with \cite{liu2020negative} and validates the dilemma exists.
Compared with the well-known overfitting between the training and testing data, such dilemma, although all validated on testing data, is more like the overfitting to base classes instead of samples. Therefore, we term it as \textbf{class-level overfitting (CO)}.
Additionally, the balance is reached when no margin is added, i.e., FSCIL cannot be improved by simply applying the margin.
For such dilemma, \cite{liu2020negative} gave an explanation by the degraded mapping from novel to base classes.
However, it could hardly be used to develop methods for handling such dilemma.
In this section, we go a step further to explain this phenomenon from the aspect of pattern learning for developing methods to handle it.

A pattern denotes a part of information that the model extracts from the input, which is a finer-grained level of analyzing the model's behavior. As studied in the interpretability of deep nets~\cite{zhou2016learning,bau2017network}, each channel in the feature extracted by deep networks could correspond to a certain pattern of the input\textsuperscript{\ref{foot:appendix}}, which can be viewed as to compose the base and novel classes~\cite{zou2020compositional}. Therefore, we conduct experiments on feature channels to study the patterns learned by applying different margins.


\vspace{-0.2cm}
\subsubsection{Class-Level Overfitting Interpreted by Pattern Fitness to Base Classes}

\begin{figure}[t]
	\centering
	\includegraphics[width=0.32\linewidth]{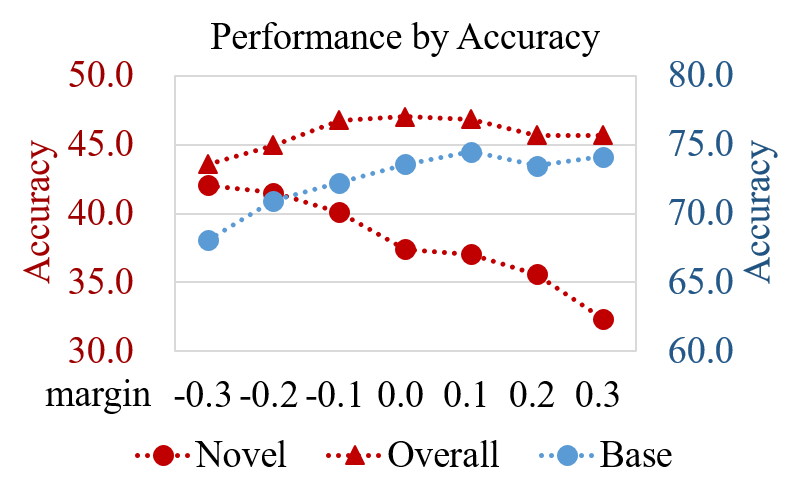}
	\centering
	\includegraphics[width=0.32\linewidth]{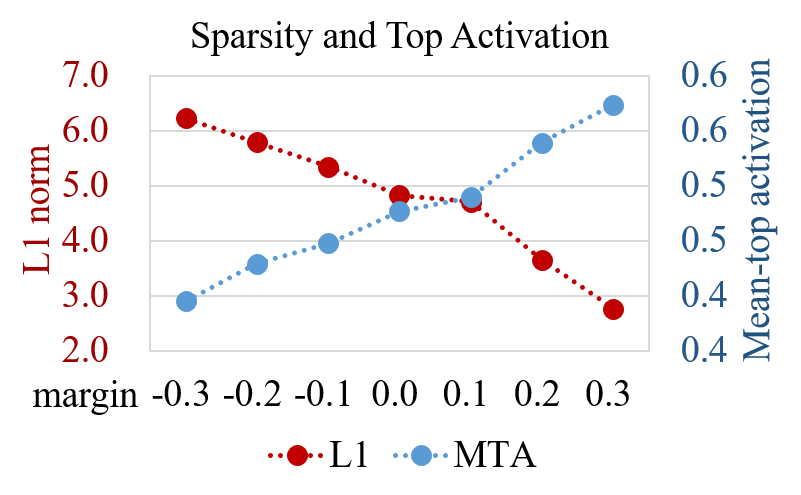}
	\centering
	\includegraphics[width=0.32\linewidth]{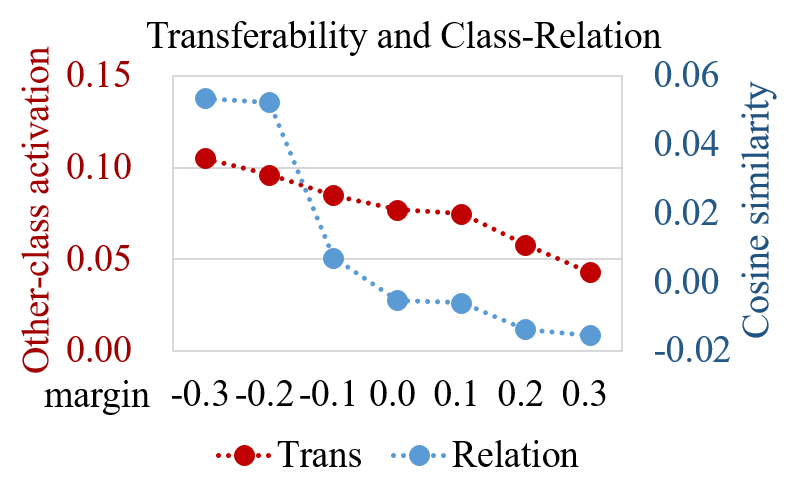}\vspace{-0.2cm}
	\caption{\textbf{Left}: Class-level overfitting exists between base and novel classes, and simply applying margins to the training can not help the overall performance. \textbf{Mid}: Pattern fits base classes more as the margin increases, making it more discriminative but less transferable. \textbf{Right}: Transferability of patterns decreases as the margin increases, pushing classes away from each other.}\vspace{-0.5cm}
	\label{fig:analysis_of_baseline}
\end{figure}

\vspace{-0.2cm}
\paragraph{Pattern's fitness to each base class.}
We first evaluate the sparsity of the base-class patterns, which is measured by the $L_1$ norm of each feature vector. As the extracted features are $L_2$ normalized, the smaller the $L_1$ norm is, for each feature, the sparser the patterns with high activation are.
Results are plotted in Fig.~\ref{fig:analysis_of_baseline} (mid), where we can see a consistent decrease in $L_1$ when the margin increases, which means the model needs less activated patterns (channels) to represent each base class. As the number of activated patterns decreases, the effectiveness of each activated pattern must increase to account for the performance increase in the base-class pre-training in Fig.~\ref{fig:analysis_of_baseline} (left). Therefore, we hold that as the margin increases, the patterns learned by the model could fit each base class better.


\vspace{-0.2cm}
\paragraph{Pattern fitness measured by the template-matching score.}
To further verify the fitness increase, we view each pattern as a semantic template and measure its matching score to each base class.
As analyzed in \cite{zhou2016learning} and \cite{bau2017network}, each pattern can be understood as a template~\cite{chen2020addernet} for the model to match the input (so that each class would has its own set of templates for recognition), and the activation can be viewed as the matching score. 
Therefore, we could know how much all patterns fit (match) each class by finding the most important patterns for each class and compare their activation.
As analyzed in \cite{zhou2016learning} and \cite{zou2020compositional}, patterns (channels) with higher weights in the classification layer are more important, and the most important ones dominate the model decisions. 
Therefore, given an input, we select its most important patterns by the top classification weights of its ground-truth class, and record the average activation on these patterns. The Mean value of such Top Activation across all samples is denoted as MTA in Fig.~\ref{fig:analysis_of_baseline} (mid).
As can be seen, as the margin increases, MTA increases consistently, which further verifies patterns' increase in fitting each base class.


\vspace{-0.2cm}
\paragraph{Better pattern fitness, worse pattern transferability.}
As each pattern could fit a corresponding base class better, its discriminability increases accordingly, but could it be transferred across classes?
To answer it, we test the transferability of patterns.
Since classes are related (e.g., cat and tiger), transferable patterns activated in one class could also be activated in other classes (e.g., felid patterns).
Therefore, we first find important patterns for each base class by the classification weights, then record activation of these patterns on \textbf{other} classes, and measure the transferability of patterns by the mean value of such other-class-activation. The results are plotted in Fig.~\ref{fig:analysis_of_baseline} (right). As can be seen, the transferability consistently decreases when the margin increases.
Combine this result with Fig.~\ref{fig:analysis_of_baseline} (mid), we hold that patterns tend to be less transferable when they fit each base class better.

\vspace{-0.2cm}
\paragraph{Discussion.}
The fitness also reflects the how much the given pattern is specific to a base class.
Imagine the extreme situation where each base-class only needs one pattern for representation, the fitness would reach its upper bound to make such pattern thoroughly specific to the corresponding class.
Therefore, we interpret that the higher the margin is, the more specific (overfitting) the patterns are to each base class, which makes patterns more discriminative but less transferable. Meanwhile, the lower the margin is, the more the patterns could be shared between classes (underfitting), making patterns more transferable but less discriminative.
The CO dilemma lies in that patterns can hardly be both class-specific and shared among classes by simply applying the classification margin.



\vspace{-0.2cm}
\subsubsection{Inherent Class Relations Lead to the Change in Pattern's Base-Class Fitness}

\vspace{-0.2cm}
\paragraph{Pattern's fitness negatively influences class relations.}

In Fig.~\ref{fig:analysis_of_baseline} (right), we also plot the class relations w.r.t. the margins. The class relations are measured by the average of cosine similarities between every two classes' prototypes. As can be seen, the relation drops as the margin grows, in consistent with the trend of the patterns' transferability. This is rationale because the if two prototypes share some patterns, the activation of the corresponding channels will be similar, making the cosine similarity larger.
As the transferability of patterns is negatively related to pattern's base-class fitness, we hold that the class relations are also negatively related to the base-class fitness.

\begin{figure}[t]
	\centering
	\begin{minipage}{0.54\textwidth}
	\centering
	\includegraphics[width=0.51\linewidth]{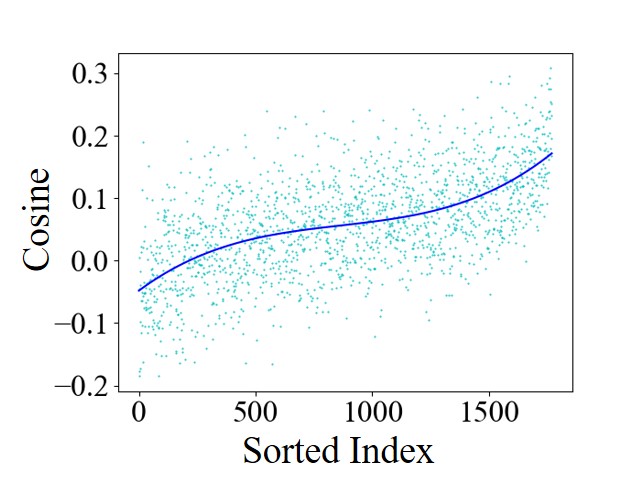}\hspace{-0.3cm}
	\centering
	\includegraphics[width=0.51\linewidth]{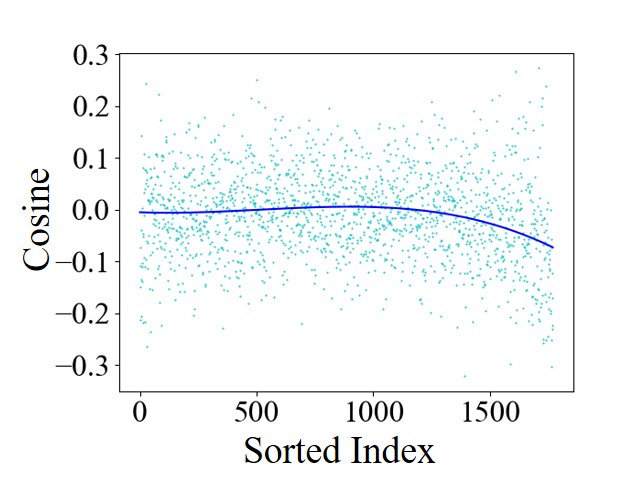}\vspace{-0.3cm}
	\caption{The change of class relations sorted by the inherent class relations. Left: negative margin. Right: positive margin.}
	\label{fig:relation_sort_vs_0}
	\end{minipage}
	\hspace{0.1cm}
	\begin{minipage}{0.44\textwidth}
	\centering
	\includegraphics[width=1.0\linewidth]{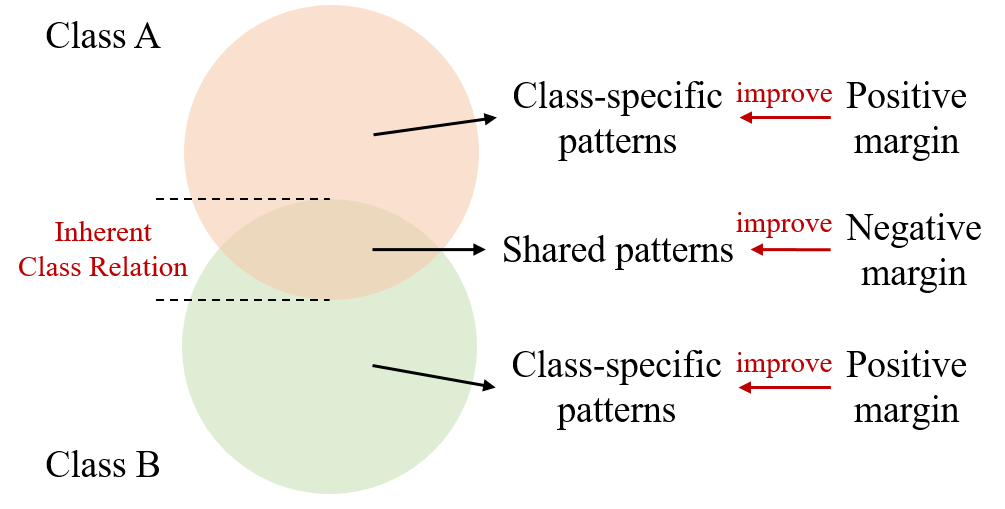}\vspace{-0.2cm}
	\caption{Intuitive interpretation of the pattern learning and inherent class relation.}
	\label{fig:explanation}	
	\end{minipage}\vspace{-0.4cm}
\end{figure}

\vspace{-0.2cm}
\paragraph{Inherent class relations influence pattern's fitness.}
The margin applied to the classification directly modifies the decision boundary between every two classes, and the decision boundary is related to the relationship between every two classes.
Therefore, we study how the class relation influences the pattern's fitness to base classes.
Specifically, given 60 base classes, for the model trained without margins, we first calculate the cosine similarity between every two different classes, which gives 60 $\times$ (60 - 1) / 2 = 1,750 relations denoted as $R_0$, which represents the \textbf{inherent relations} between all classes. Similarly, we calculate 1,750 relations for the model trained with positive and negative margins respectively, denoted as $R_{pos}$ and $R_{neg}$. Then we calculate $D_{pos} = R_{pos} - R_0$ and $D_{neg} = R_{neg} - R_0$. Finally, we sort $R_0$ from small to large, and use the sorted index to arrange $D_{neg}$ and $D_{pos}$. As plotted in Fig.~\ref{fig:relation_sort_vs_0}, the blue dots denote $D_{neg}$ (left) or $D_{pos}$ (right), and the blue curve are dots fitted by multinomial functions. We can see that the change in class relations is positively related to the inherent class relations for the negative margins, while is negatively related to the inherent class relations for the positive margins, especially for index larger than 1,000. 

Since class relations are negatively related to pattern's fitness to base classes, the more the class relation increases, the less the patterns could fit the corresponding base class. Therefore, the results can be understood that the more inherently similar two classes are, by applying a negative margin, the less the patterns could fit the given base class, i.e., the more the patterns are shared by given classes, making these classes' representations more similar; by applying a positive margin, the more the patterns could fit the given base classes, making these classes' representations more dissimilar.

\vspace{-0.2cm}
\paragraph{Conclusion and Discussion.}
Therefore, as shown in Fig.~\ref{fig:explanation}, we interpret the pattern learning process as follows. 
Given a negative margin, the decision boundary between two classes are confused, making more samples fall into the overlapping region between two classes. This makes it possible to learn patterns shared by these two classes and hinder the pattern from fitting the given base class, and makes patterns more transferable but less discriminative. The more similar inherently two classes are (i.e., larger overlapping region in Fig.~\ref{fig:explanation}), the more shared patterns can be learned (e.g., more patterns could be shared between cats and tigers than cats and air-planes), therefore making these classes' representations more similar.
On the contrary, given a positive margin, the decision boundary between two classes should be well separated, pushing the model to learn patterns fitting (i.e., specific to) each class, which are more discriminative but less transferable. The more similar inherently two classes are, the harder the learning is and the larger the training loss will be, therefore making the patterns fit each class more and making these classes' representations more dissimilar. 


\vspace{-0.3cm}
\section{Mitigating the Dilemma of Few-Shot Class-Incremental Learning}
\label{sec:method}
\vspace{-0.2cm}

\begin{figure}[t]
	\centering
	\includegraphics[width=0.9\linewidth]{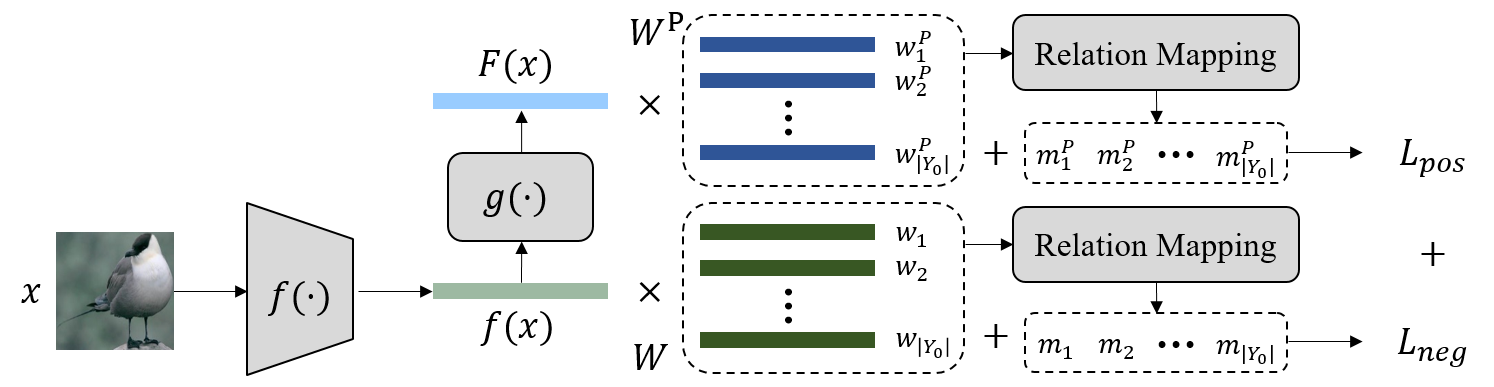}\vspace{-0.2cm}
	\caption{Method (CLOM) framework. We construct the positive-margin-based (PM, $F(\cdot)$) feature from the negative-margin-based (NM, $f(\cdot)$) feature, and map class relations to form a set of class-specific margins, which can effectively mitigate the dilemma of class-level overfitting by providing extra constraint to the NM/PM pattern learning through the learning of PM/NM patterns respectively.}\vspace{-0.4cm}
	\label{fig:framework}
\end{figure}

In this section, we first analyze the cause of the dilemma in margin-based classification based on the above interpretation, then we propose our method (named as Class-Level Overfitting Mitigation, CLOM) to mitigate the CO dilemma based on the analysis, as shown in Fig.~\ref{fig:framework}.

\vspace{-0.2cm}
\subsection{Analysis on the Cause of Class-Level Overfitting}
\vspace{-0.1cm}

Based on the above analysis,
we can find the learning of negative-margin-based patterns loosely constrains the given pattern to be shared by both classes. However, such constraint is easily satisfied by ineffective patterns as simple as edges or corners, which could lead to the low discriminability of negative-margin-based patterns. Similarly, the learning of positive-margin-based patterns loosely constrains the given pattern to be specific to the given class. However, such constraint could be satisfied by easily finding patterns sharing no information with other classes, such as finding some complex texture only specific to the given class, which could lead to the low transferability of positive-margin-based patterns. 
Such \textbf{easy-constraint} problems push patterns to be \textbf{only} class-specific or shared among classes, making pattern effective in one scenario ineffective in other ones. 

To verify the above claims, we analyze how simple or complex the learned patterns are. 
Inspired by \cite{kornblith2019similarity}, we quantitatively measure the simplicity/complexity of patterns by the similarity between the extracted feature and the simplest feature (e.g., corner or edge features).

We first use the baseline model to train on CIFAR100, and use the first convolutional layer as the simplest feature extractor (denoted as $f_{simple}$), since many works (e.g., \cite{yosinski2015understanding}) has shown that the first convolutional layer tends to capture corners or edges. Then, we train models with different margins, and use the backbone network for feature extraction (denoted as $f_{target}$). After that, we extract $f_{simple}$ and $f_{target}$ features from all images in base classes. Finally, we compare the CKA similarity~\cite{kornblith2019similarity} for measuring the similarity between $f_{simple}$ and the $f_{target}$.

For a sanity check, we first report the similarity between different layers within the baseline model.

\vspace{-0.4cm}
\begin{table}[h]
	\caption{Sanity check for the CKA measure.}
	\label{tab:CKA_sanity}
	\centering
	\resizebox{0.88\textwidth}{!}{\begin{tabular}{c|ccccc}
			\toprule
			$f_{target}$ & Conv1-output &  Stage1-output &  Stage2-output &  second-last-Conv &  backbone-output \\
			\midrule
			CKA 		 & 1.0  		&  0.8876 		 &  0.5664   	  &  0.2097     	  &  0.1306 \\
			\bottomrule
	\end{tabular}}\vspace{-0.2cm}
\end{table}

We can see that the shallower the layer is, the higher the CKA similarity would be, which means the more similar they are to the $f_{simple}$, i.e. the patterns are simpler, more transferable but less discriminative~\cite{yosinski2014transferable}. Then, we report the comparison of the CKA similarity between $f_{simple}$ and $f_{target}$ (backbone feature) of the baseline model trained with margins.

\vspace{-0.4cm}
\begin{table}[h]
	\caption{CKA between the $f_{simple}$ and baseline backbone features trained with different margins.}
	\label{tab:CKA_baseline}
	\centering
	\resizebox{0.95\textwidth}{!}{\begin{tabular}{c|cccccccccc}
			\toprule
			Margin &	-0.5  &  -0.4  &  -0.3  &  -0.2  &  -0.1  &   0.0  &  0.1   &  0.2   &  0.3   &  0.4   \\
			\midrule
			CKA    &  0.2432  & 0.2245 & 0.2010 & 0.1661 & 0.1510 & 0.1306 & 0.1149 & 0.0837 & 0.0642 & 0.0576 \\ 
			\bottomrule
	\end{tabular}}\vspace{-0.2cm}
\end{table}

We can see that by applying a negative margin, the CKA similarity clearly increases (even larger than that of the second last convolutional layer when margin < -0.3), showing that the backbone network captures patterns more similar to simplest ones such as edges or corners, which verifies that the model tends to learn simple patterns that are easily shared between classes. When applying a positive margin, the captured patterns grow to be more complex and tend to overfit base classes, making the CKA much smaller than baseline model's backbone-output, which verifies that the model tends to learn complex patterns that are easily specific to a given base class.

Therefore, the key to mitigating the CO dilemma is to extra constrain the pattern learning process.
%

\vspace{-0.2cm}
\subsection{Mitigating Class-Level Overfitting by Providing Extra Constraint}
\vspace{-0.1cm}

Negative-Margin-based (NM) patterns are more transferable and class-shared, while Positive-Margin-based (PM) patterns are discriminative and class-specific. These characteristics are similar to the behavior of features from the shallow and deep layers of deep networks~\cite{yosinski2014transferable,zou2021revisiting}. Generally, shallow-layer features encode low-level patterns that are easily shared by most objects therefore more transferable, such as edge or corner, while the deep-layer features encode high-level patterns that are semantically related to only few object classes therefore more discriminative. Such similarity inspires us to view PM patterns as high-level patterns while view NM patterns as (relatively) low-level patterns, and build PM patterns from the NM patterns, just like building high-level features from low-level features in deep networks. As the learning of NM patterns is influenced by PM patterns, this design could benefit the learning of NM patterns by providing extra constraint from the learning of PM patterns, and vice versa, which could therefore handle the easy-constraint problem.

Specifically, given an feature extractor $f(\cdot)$, we add one more layer to construct the PM feature as 
\vspace{-0.1cm}
\begin{equation}
	F(x_i) = g(f(x_i)),
	\label{eq:F}
\end{equation}
where $F(\cdot)$ is the PM feature extractor, $g(\cdot)$ is typically composed of a fully-connected layer, a batch-normalization layer~\cite{ioffe2015batch} and an activation layer. During training, two classification loss is applied to both $F(\cdot)$ and $f(\cdot)$ with positive and negative margins respectively, represented as
\vspace{-0.1cm}
{\small
\begin{align}
&	L(x_i, y_i) = L_{neg}(x_i, y_i) + L_{pos}(x_i, y_i) \\	
&	= -log \frac{e^{\tau (w_{y_i} f(x_i) - m_{neg})}}{e^{\tau (w_{y_i} f(x_i) - m_{neg})} + \sum_{j \neq y_i} e^{\tau w_j f(x_i)}} - \lambda \cdot log \frac{e^{\tau (w^P_{y_i} F(x_i) - m_{pos})}}{e^{\tau (w^P_{y_i} F(x_i) - m_{pos})} + \sum_{j \neq y_i} e^{\tau w^P_j F(x_i)}},
	\label{eq:ind_loss}
\end{align}
}
\hspace{-0.1cm}where $w^P_j \in R^{1 \times d^P}$ denotes the classifier weights for $F(\cdot)$ corresponding to class $j$, $m_{neg}$ denotes the negative margin for $f(\cdot)$, $m_{pos}$ is the positive margin for $F(\cdot)$, and $\lambda$ is typically set to 1.0.

Given this modification, the NM patterns $f(\cdot)$ are utilized to construct the discriminative PM patterns $F(\cdot)$. Since $g(\cdot)$ is relatively too simple to capture complex patterns, $f(\cdot)$ will be pushed to be discriminative (validated in Fig.~\ref{fig:f_neg}), instead of casually learning ineffective patterns shared between similar classes. Similarly, as $F(\cdot)$ are built from the transferable $f(\cdot)$ by the simple $g(\cdot)$, the transferability could be maintained by pushing PM patterns to fit the corresponding base class by transferable information, which improves the transferability of $F(\cdot)$ (validated in Fig.~\ref{fig:f_pos}).

During testing, given the improved $f(\cdot)$ and $F(\cdot)$, the final feature would be their concatenation.

\vspace{-0.2cm}
\subsection{Further Mitigation of Class-Level Overfitting by Integrating Class Relations}
\vspace{-0.1cm}

Based on the architecture design, we propose to further mitigate the CO problem through boosting the discriminability and tranferability of NM and PM patterns by integrating the class relations.

For easy understanding, we first introduce the modification to the classification for $f(\cdot)$, and later introduce that for $F(\cdot)$. To modify the margin-based classification from a single margin to margins related to class relations, we first move the margin from the ground-truth logit to all other logits as
\vspace{-0.15cm}
\begin{equation}
	L(x_i, y_i) = -log \frac{e^{\tau w_{y_i} f(x_i)}}{e^{\tau w_{y_i} f(x_i)} + \sum_{j \neq y_i} e^{\tau (w_j f(x_i) + m(A_{ij}))}},
	\label{eq:adj_m}
\end{equation}
where $A$ is the adjacency matrix between all classes, $m(\cdot)$ maps the adjacency value between two classes to a margin. Given this modification, if the margin is set to the original fixed value as Eq.~\ref{eq:margin}, the decision boundary between class $y_i$ and other classes remains the same, therefore the new loss in Eq.~\ref{eq:adj_m} could be an approximation of the original loss in Eq.~\ref{eq:margin}. 

For measuring class relations, we choose to utilize the adjacency matrix of all classes. As both the feature and the classifier's weights are $L_2$ normalized, the adjacency can be measured by the cosine similarity between every two class prototypes as $A_{ij} = cos(P_i, P_j)$, where $P_i$ is the prototype for class $i$, which is typically set to $w_i$ of the classifier.

If two classes are identical, the cosine similarity would reach its upper bound to 1.0. Meanwhile, the average of class relations reflects a global margin that is effective for most classes. Therefore, we design to interpolate the margin from a global effective value to a pre-defined upper bound value as
\vspace{-0.15cm}
\begin{equation}
	m(A_{ij}) = m_{ave} + \frac{m_{upper} - m_{ave}}{1.0 - A_{ave}} \cdot (A_{ij} - A_{ave}),
	\label{eq:map_from_adj_to_margin}
\end{equation}
where $m_{upper}$ is a hyper-parameter controlling the margin for the upper-bound class relation (i.e., two identical classes), $m_{ave}$ is another hyper-parameter controlling the margin for the average class relations, set to the same value as that in Eq.~\ref{eq:margin}, and $A_{ave}$ is the average class relations calculated as $A_{ave} = \frac{1}{|Y_0| \cdot (|Y_0| - 1)} \sum_{j=1}^{j=|Y_0|} \sum_{k \neq j} A_{jk} $.

For $F(\cdot)$, we adopt the same modification as $f(\cdot)$, which replaces $f(\cdot)$ with $F(\cdot)$ in Eq.~\ref{eq:adj_m}, and replaces $A$ with $A^P$ in Eq.~\ref{eq:adj_m} where $A^P_{ij} = cos(w^P_i, w^P_j)$.
Therefore, the final training objective is
\vspace{-0.15cm}
\begin{flalign}\hspace{-0.35cm}
	\resizebox{0.97\textwidth}{!}{
		$L(x_i, y_i) = - log \frac{e^{\tau w_{y_i} f(x_i)}}{e^{\tau w_{y_i} f(x_i)} + \sum\limits_{j \neq y_i} e^{\tau (w_j f(x_i) + m_{n}(A_{ij}))}}
		- \lambda \cdot log \frac{e^{\tau w^P_{y_i} F(x_i)}}{e^{\tau w^P_{y_i} F(x_i)} + \sum\limits_{j \neq y_i} e^{\tau (w^P_j F(x_i) + m_{p}(A^P_{ij}))}}$,
	}
	\label{eq:final_loss}
\end{flalign}
where $m_{p}(\cdot)$ and $m_{n}(\cdot)$ are two interpolate functions as Eq.~\ref{eq:map_from_adj_to_margin} with positive and negative hyper-parameters ($m^P_{upper}$, $m^P_{ave}$) and ($m_{upper}$, $m_{ave}$) respectively.

In experiments (Fig.~\ref{fig:margin_by_adj}), we find it beneficial to have $m_{upper} < m_{ave}$ and $m^P_{upper} > m^P_{ave}$, which means for classes with higher similarities, the relation mapping module enables the model to learn more shared (transferable) patterns by applying smaller negative margins, and learn more class-specific (discriminative) patterns by applying larger positive margins.
Moreover, due to the connection between NM and PM patterns, the improved transferability in NM patterns would be transmitted to PM patterns and vice versa, as validated in Fig.~\ref{fig:margin_by_adj}, which further mitigates the CO dilemma.

After the base-session training, the feature extractor $F(\cdot)$ and $f(\cdot)$ will be applied to the training data of each session to obtain the extended classifier weight $W = \{w^0_1, w^0_2, ..., w^0_{|Y_0|}\} \cup ... \cup \{w^k_1, ..., w^k_{|Y_k|}\}$ by averaging extracted features, and the final performance of each session will be obtained based on the classification of all the encountered classes' test samples.

\begin{table}[t]
	\caption{Evaluation datasets.}
	\label{tab:dataset}
	\centering
	\resizebox{1.0\textwidth}{!}{\begin{tabular}{lcccccc}
			\toprule
			Dataset & Total Classes & Base Classes & Novel Classes & Incremental Sessions & Novel-Class Shot & Input Size \\
			\midrule
			CIFAR100 & 100 & 60 & 40 & 8 & 5 & 32 $\times$ 32 \\
			CUB200 & 200 & 100 & 100 & 10 & 5 & 224 $\times$ 224 \\
			\textit{mini}ImageNet & 100 & 60 & 40 & 8 & 5 & 84 $\times$ 84 \\
			\bottomrule
	\end{tabular}}\vspace{-0.5cm}
\end{table}

\begin{table}[t]
	\caption{Comparison with state-of-the-art works on the CUB200 dataset.}
	\label{tab:cub_sota}
	\centering
	\resizebox{1.0\textwidth}{!}{\begin{tabular}{lccccccccccc}
			\toprule
			Method & S0 & S1 & S2 & S3 & S4 & S5 & S6 & S7 & S8 & S9 & S10 \\
			\midrule
			Finetune & 68.68 & 43.70 & 25.05 & 17.72 & 18.08 & 16.95 & 15.10 & 10.06 & 8.93 & 8.93 & 8.47 \\
			Rebalancing~\cite{hou2019learning} & 68.68 & 57.12 & 44.21 & 28.78 & 26.71 & 25.66 & 24.62 & 21.52 & 20.12 & 20.06 & 19.87 \\
			iCaRL~\cite{rebuffi2017icarl} & 68.68 & 52.65 & 48.61 & 44.16 & 36.62 & 29.52 & 27.83 & 26.26 & 24.01 & 23.89 & 21.16 \\
			EEIL~\cite{castro2018end} & 68.68 & 53.63 & 47.91 & 44.20 & 36.30 & 27.46 & 25.93 & 24.70 & 23.95 & 24.13 & 22.11 \\
			TOPIC~\cite{tao2020few} & 68.68 & 62.49 & 54.81 & 49.99 & 45.25 & 41.40 & 38.35 & 35.36 & 32.22 & 28.31 & 26.26 \\
			Decoupled-NegCosine~\cite{liu2020negative} & 74.96 & 70.57 & 66.62 & 61.32 & 60.09 & 56.06 & 55.03 & 52.78 & 51.50 & 50.08 & 48.47 \\
			CEC~\cite{zhang2021few} & 75.85 & 71.94 & 68.50 & 63.50 & 62.43 & 58.27 & 57.73 & 55.81 & 54.83 & 53.52 & 52.28 \\
			FSLL+SS~\cite{mazumder2021few} & 75.63 & 71.81 & 68.16 & 64.32 & 62.61 & 60.10 & 58.82 & 58.70 & 56.45 & 56.41 & 55.82 \\
			FACT~\cite{zhou2022forward} & 75.90 & 73.23 & 70.84 & 66.13 & 65.56 & 62.15 & 61.74 & 59.83 & 58.41 & 57.89 & 56.94 \\
			IDLVQ-C~\cite{chen2020incremental} & 77.37 & 74.72 & 70.28 & 67.13 & 65.34 & 63.52 & 62.10 & 61.54 & 59.04 & 58.68 & 57.81 \\
			\midrule
			CLOM (Ours) & \textbf{79.57} & \textbf{76.07} & \textbf{72.94} & \textbf{69.82} & \textbf{67.80} & \textbf{65.56} & \textbf{63.94} & \textbf{62.59} & \textbf{60.62} & \textbf{60.34} & \textbf{59.58} \\
			\bottomrule
	\end{tabular}}\vspace{-0.4cm}
\end{table}

\vspace{-0.3cm}
\section{Experiments}
\label{sec:exp}
\vspace{-0.2cm}

\begin{figure}[t]
	\centering
	\includegraphics[width=0.45\linewidth]{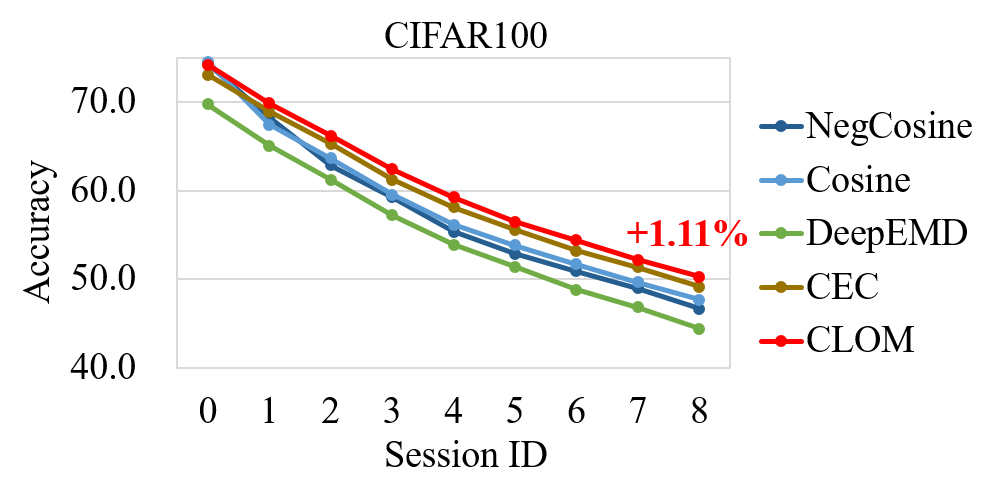}
	\centering
	\includegraphics[width=0.45\linewidth]{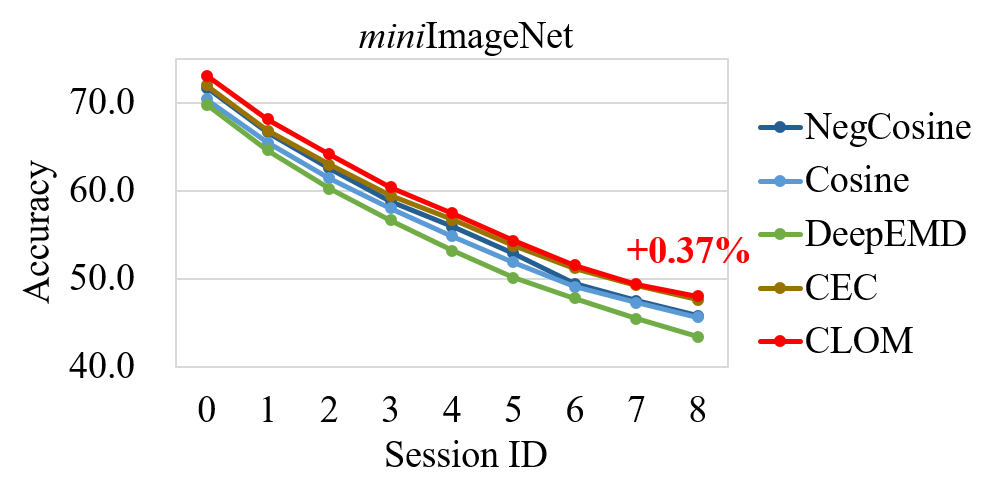}\vspace{-0.2cm}
	\caption{Comparison with state-of-the-art works on CIFAR100 and \textit{mini}ImageNet.}\vspace{-0.5cm}
	\label{fig:cifar_and_mini}
\end{figure}

In this section, we first introduce the experiment settings, then compare the proposed method with the state-of-the-art methods, and finally report the ablation study for the effectiveness of each design.

\vspace{-0.3cm}
\subsection{Datasets}
\vspace{-0.1cm}




Datasets include CIFAR100~\cite{krizhevsky2009learning}, Caltech-UCSD Birds-200-2011 (CUB200)~\cite{wah2011caltech} and \textit{mini}-ImageNet~\cite{Vinyals2016Matching} as listed in Tab.~\ref{tab:dataset} following the split in \cite{tao2020few}. For details, please refer to the appendix.

\vspace{-0.3cm}
\subsection{Implementation Details}
\vspace{-0.1cm}

The implementation is based on CEC's code~\cite{zhang2021few}, and our code will be released\footnote{https://github.com/Zoilsen/CLOM}. For CIFAR100, we set $d^P$=256, set $m_{ave}$=-0.2, set $m_{upper}$=-0.5, and we have $m^P_{ave}$=0.1 and $m^P_{upper}$=0.2. For CUB200, we scale the learning rate of the backbone network to 10\% of the global learning rate since the pre-training of the backbone is adopted~\cite{zhou2022forward,zhang2021few}, and set $d^P$ to 8192. Then we have $m_{ave}$=-0.2 and $m_{upper}$=-0.25 and $m^P_{ave}$=0.3 and $m^P_{upper}$=0.6. For \textit{mini}ImageNet, we set $d^P$ to 4096, and have $m_{ave}$=-0.2 and $m_{upper}$=-0.5 and $m^P_{ave}$=0.1 and $m^P_{upper}$=0.2. 
Please refer to appendix for details.

\begin{table}[t]
	\begin{center}
		\caption{Ablation study of modules on the last incremental session of three datasets. }
		\label{tab:ablation}
		\resizebox{0.8\textwidth}{!}{
			\begin{tabular}{lccccccccc}
				\toprule
				\multirow{2}{*}{\tabincell{c}{Method}} & \multicolumn{3}{c}{CUB200} & \multicolumn{3}{c}{CIFAR100} & \multicolumn{3}{c}{\textit{mini}ImageNet} \\
				\cmidrule{2-10}
				& Overall & Novel & Base & Overall & Novel & Base & Overall & Novel & Base \\
				\midrule
				Baseline    & 57.78 & 45.97 & 79.48 & 47.02 & 37.40 & 72.32 & 46.58 & 31.02 & 72.33 \\
				+ $g(\cdot)$& 57.21 & 47.13 & 79.39 & 48.37 & 39.55 & 72.70 & 46.79 & 30.97 & 72.60 \\
				+ Margin    & 58.73 & 49.93 & 79.47 & 49.21 & 40.22 & 73.72 & 47.30 & 32.07 & 72.93 \\
				+ Relation  & \textbf{59.58} & \textbf{50.89} & \textbf{79.57} & \textbf{50.25} & \textbf{41.17} & \textbf{74.20} & \textbf{48.00} & \textbf{33.60} & \textbf{73.08} \\
				\bottomrule
		\end{tabular}}\vspace{-0.6cm}
	\end{center}
\end{table}

\vspace{-0.3cm}
\subsection{Comparison with the State-of-the-Art}
\vspace{-0.1cm}

Comparisons with the state-of-the-art works are listed in Tab.~\ref{tab:cub_sota} and Fig.~\ref{fig:cifar_and_mini}, where we can achieve state-of-the-art performance on all three datasets.
Specifically, we can first see that our method, as a prototype-based method (e.g., CEC~\cite{zhang2021few}), could significantly outperform finetune-based methods (e.g., iCaRL~\cite{rebuffi2017icarl}). This is because the few-shot training data could not provide sufficient information for novel-class learning, therefore directly freezing parameters on novel classes could reduce the catastrophic-forgetting problem brought by the finetuning. Our method also outperforms other prototype-based methods, this is because the core of the prototype-based methods lies in the metric learning~\cite{snell2017prototypical}. As empirically proved by current works~\cite{sun2020circle,wang2018cosface,deng2019arcface,liu2020negative}, applying the margin-based classification could effectively improve the embedding space learned by metric-based methods. Since our method handles the difficulty of applying margin-based classification (i.e., class-level overfitting) to the FSCIL task, our method could beat these prototype-based ones. For detailed numbers of CIFAR100 and \textit{mini}ImageNet, please refer to the appendix.

\vspace{-0.3cm}
\subsection{Ablation Study of Modules}
\vspace{-0.1cm}

The ablation study of each module is reported in Tab.~\ref{tab:ablation}, where $g(\cdot)$ denotes adding another layer, \textit{Margin} denotes the applicant of margin-based classification, and \textit{Relation} refers to the relation mapping of the margin. We study from three aspects: \textit{Base}-class, \textit{Novel}-class, and \textit{Overall} accuracy of the last incremental session.
From Tab.~\ref{tab:ablation}, we can see that 

$\bullet$ \textit{Simply applying another layer cannot consistently improve the performance}.

$\bullet$ \textit{The designed architecture could mitigate class-level overfitting}. Compared with experiments in Fig.~\ref{fig:analysis_of_baseline} where no \textit{Overall} performance improvements can be obtained by simply adding margins, the performance here is clearly improved by adding margins on the designed architecture. Moreover, the overall improvements originate from not only the improved \textit{Base} performance, but also the boosted \textit{Novel} performance, demonstrating the mitigation of class-level overfitting (CO) problem.

$\bullet$ \textit{Relation mapping could further improve performance by mitigating class-level overfitting}.

\vspace{-0.2cm}
\subsection{Verification of Class-Level Overfitting Mitigation}
\vspace{-0.1cm}

\begin{figure}[t]
	\centering
	\includegraphics[width=0.26\linewidth]{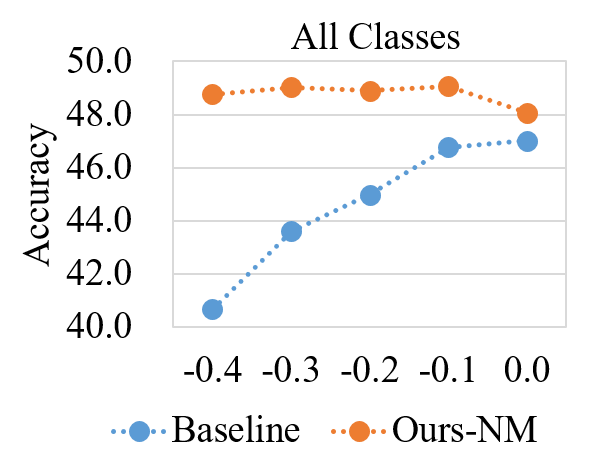}
	\centering
	\includegraphics[width=0.26\linewidth]{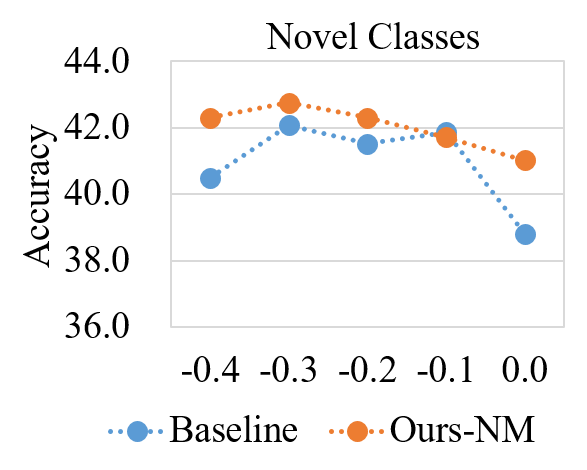}
	\centering
	\includegraphics[width=0.26\linewidth]{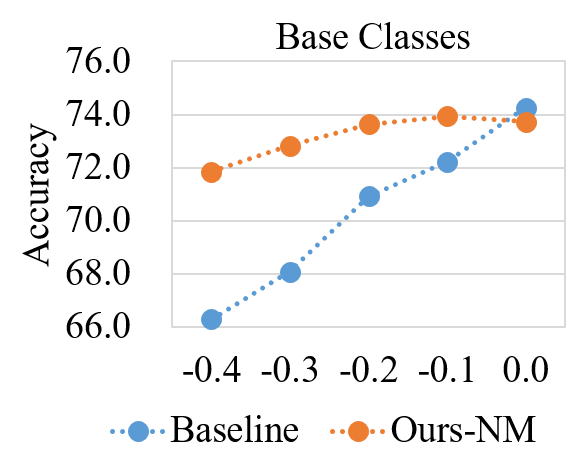}\vspace{-0.2cm}
	\caption{NM feature's accuracy compared with the baseline with negative margins (CIFAR100).}
	\label{fig:f_neg}
	\centering
	\includegraphics[width=0.26\linewidth]{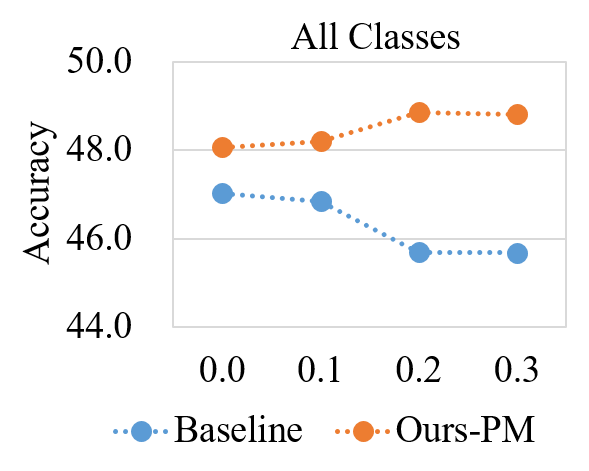}
	\centering
	\includegraphics[width=0.26\linewidth]{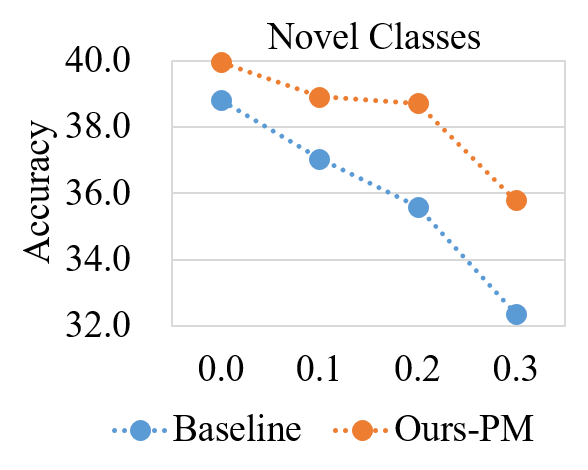}
	\centering
	\includegraphics[width=0.26\linewidth]{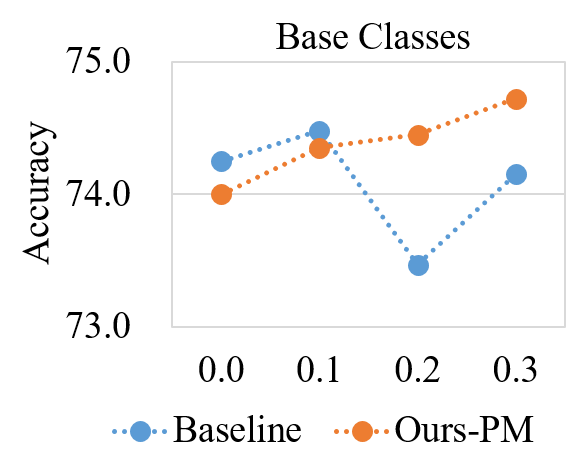}\vspace{-0.2cm}
	\caption{PM feature's accuracy compared with the baseline with positive margins (CIFAR100).}\vspace{-0.5cm}
	\label{fig:f_pos}
\end{figure}

\begin{figure}[t]
	\centering
	\includegraphics[width=0.24\linewidth]{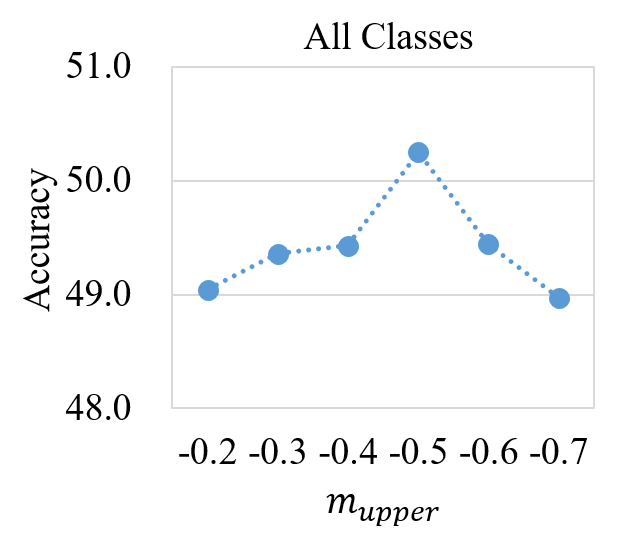}
	\centering
	\includegraphics[width=0.24\linewidth]{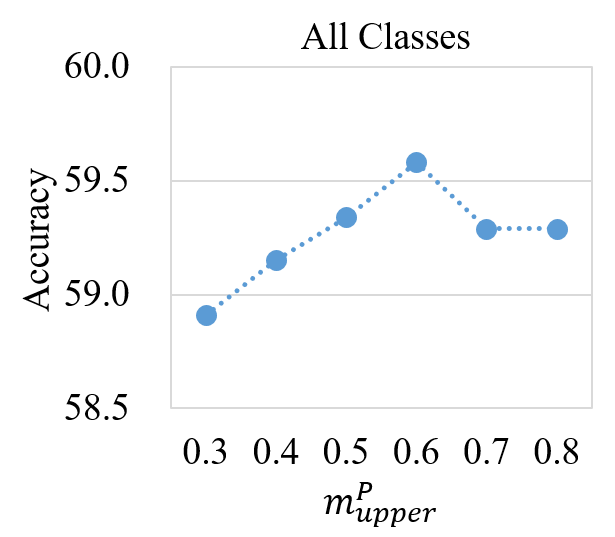}
	\centering
	\includegraphics[width=0.24\linewidth]{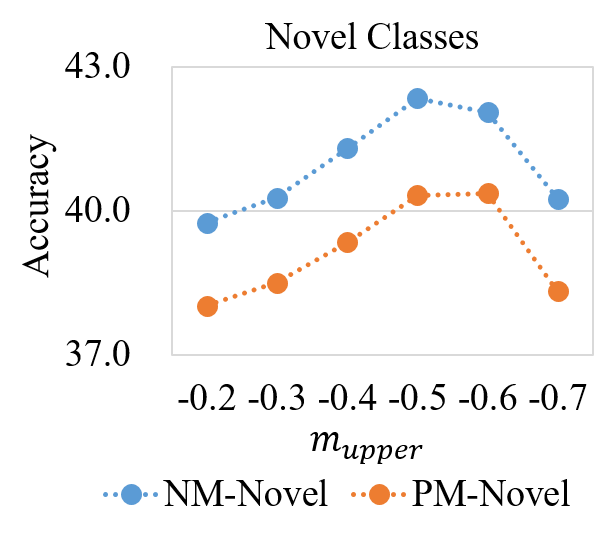}
	\centering
	\includegraphics[width=0.24\linewidth]{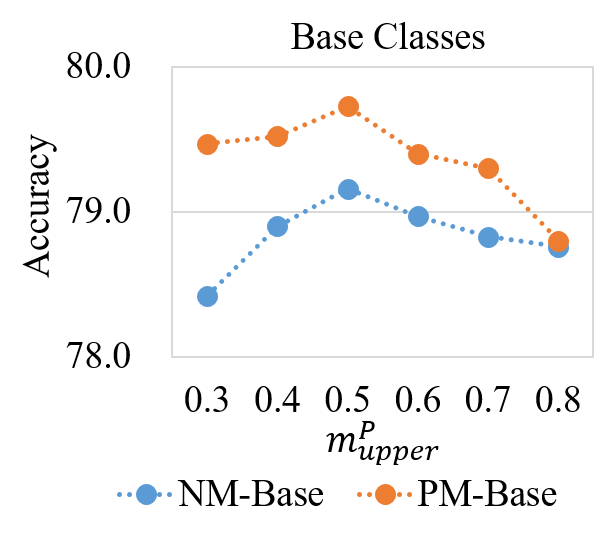}\vspace{-0.1cm}
	\caption{Upper margins in the relation mapping module ($m_{upper}$: CIFAR100, $m^P_{upper}:$ CUB200).}\vspace{-0.5cm}
	\label{fig:margin_by_adj}
\end{figure}


\vspace{-0.2cm}
\paragraph{Mitigating the class-level overfitting by the architecture design.} We first compare our method with the baseline method which applies the margin directly to the backbone feature in Fig.~\ref{fig:f_neg} and \ref{fig:f_pos}, so as to validate the mitigation of CO. In Fig.~\ref{fig:f_neg}, we apply the negative margin to the backbone (NM) feature while applying no margin to the PM feature, and see the accuracy of the NM feature against the baseline. As can be seen, while keeping the improvements on novel classes (mid), our method (orange) maintains the base-class accuracy (right), in contrast to a sharp decrease of the baseline (blue) when the margin decreases, which makes our NM feature significantly outperforms the baseline in terms of the overall accuracy (left). Similarly, in Fig.~\ref{fig:f_pos}, our PM feature also maintains a higher performance on novel classes when the margin increases. These experiments validate the mitigation of the CO dilemma: by implicitly guiding the learning of the PM and NM features by each other, the transferability or discriminability drop on them can be mitigated respectively.





\vspace{-0.3cm}
\paragraph{Further mitigating class-level overfitting by relation mapping.} To validate the relation mapping module, in Fig.~\ref{fig:margin_by_adj} we fixed the average margin (i.e., $m_{ave}$=-0.2 in Fig.~\ref{fig:margin_by_adj}.1st and $m^P_{ave}$=0.3 in Fig.~\ref{fig:margin_by_adj}.2nd) and experiment with different upper margins (i.e., $m_{upper}$ in Fig.~\ref{fig:margin_by_adj}.1st and $m^P_{upper}$ in Fig.~\ref{fig:margin_by_adj}.2nd). Similar to Fig.~\ref{fig:f_neg}, when conducting experiments on one branch, the hyper-parameters of the other branch are fixed. We can see the relation mapping module can indeed help the learning, as the upper margins are significantly different from the average margin.
We also plot the NM and PM performance on base classes with negative margins in Fig.~\ref{fig:margin_by_adj}(3rd), and those on novel classes with positive margins in Fig.~\ref{fig:margin_by_adj}(4th). We can see the NM and PM performance are improved simultaneously, which validates the improvements brought by the extra guidance of each features. Therefore, the relation mapping module can also help to mitigate the class-level overfitting problem by building PM features with more transferable patterns, and vice versa.
Moreover, it is interesting to find the performance is improved when $m_{upper} < m_{ave}$ and $m^P_{upper} > m^P_{ave}$, which means for classes with higher similarities, the relation mapping module enables the model to learn more shared patterns by applying negative margins with larger absolute value, and learn more class-specific patterns by applying larger positive margins. Also, this is consistent with the results in Fig.~\ref{fig:relation_sort_vs_0}.

\vspace{-0.2cm}
\subsection{Mitigating the Easy-Constraint Problem}
\vspace{-0.1cm}

Following experiments in Tab.~\ref{tab:CKA_sanity}, we also utilize the CKA similarity between the simplest feature $f_{simple}$ and the NM/PM feature to validate the mitigation of the easy-constraint problem.
By applying our method, for the NM (or PM) patterns, we set $f_{target}$ to $f(\cdot)$ (or $F(\cdot)$) in Fig.5 and set the positive/negative margin to 0.3/-0.4, and report the CKA below.

\begin{minipage}{\textwidth}	
	\tabcaption{CKA between $f_{simple}$ and NM (left) or PM (right) features trained with different margins.}\vspace{-0.2cm}
	\label{tab:CKA_NM+PM}
	\begin{minipage}[t]{0.53\textwidth}		
		\resizebox{1.0\textwidth}{!}{
			\begin{tabular}{c|cccccc}
				\toprule
				Margin 	 & -0.5   &  -0.4   &   -0.3  &  -0.2  &   -0.1  &  0.0   \\
				\midrule 
				CKA 	 & 0.1867 &  0.1779 &  0.1724 & 0.1638 &  0.1552 & 0.1427 \\
				\bottomrule
			\end{tabular}
		}
	\end{minipage}
	\begin{minipage}[t]{0.45\textwidth}
		\resizebox{1.0\textwidth}{!}{
			\begin{tabular}{c|ccccc}
				\toprule
				Margin	 &   0.0  &   0.1   &   0.2   &   0.3   &   0.4   \\
				\midrule
				CKA	     & 0.1476 &  0.1439 &  0.1430 &  0.1214 &  0.1100 \\
				\bottomrule
			\end{tabular}
		}
	\end{minipage}
\end{minipage}

In Tab.~\ref{tab:CKA_NM+PM} (left), compared with Tab.~\ref{tab:CKA_baseline}, CKA decreases clearly compared with the baseline method, which validates that NM patterns learned by our method are more complex and less similar to edges and corners than the baseline method, verifying the mitigation of the easily-satisfied constraint problem by extra supervision from the learning of PM patterns.
Also, in Tab.~\ref{tab:CKA_NM+PM} (right), compared with Tab.~\ref{tab:CKA_baseline}, CKA is much larger than that of the baseline method, which validates that the PM patterns are more similar to the simplest patterns than the baseline method, which makes it less overfitting the base classes, verifying the mitigation by extra supervision from the learning of NM patterns.

\vspace{-0.3cm}
\section{Conclusion}
\vspace{-0.2cm}

In this paper, we focus on the dilemma in the margin-based classification for FSCIL. We first interpret the dilemma as a class-level overfitting problem from the aspect of pattern learning, then find the cause of this problem lies in the easily-satisfied constraint of learning shared or class-specific patterns. Based on the analysis, we design a method (CLOM) to mitigate the dilemma by constructing PM patterns from NM patterns, and mapping class relations into class-specific patterns. Extensive experiments on three public datasets validate the effectiveness and outstanding performance of the proposed method.

\vspace{-0.3cm}
\section*{Acknowledgements}
\vspace{-0.2cm}

This work is supported by National Natural Science Foundation of China under grants U1836204, U1936108, 62206102, Science and Technology Support Program of Hubei Province under grant 2022BAA046, and 2022 CCF-DiDi Gaiya Young Scholar Research Fund.

\bibliographystyle{plain}
\bibliography{yixiongz}

\section*{Checklist}


\begin{enumerate}

\item For all authors...
\begin{enumerate}
  \item Do the main claims made in the abstract and introduction accurately reflect the paper's contributions and scope?
    \answerYes{}
  \item Did you describe the limitations of your work?
    \answerYes{}
  \item Did you discuss any potential negative societal impacts of your work?
    \answerYes{}
  \item Have you read the ethics review guidelines and ensured that your paper conforms to them?
    \answerYes{}
\end{enumerate}

\item If you are including theoretical results...
\begin{enumerate}
  \item Did you state the full set of assumptions of all theoretical results?
    \answerNA{}
        \item Did you include complete proofs of all theoretical results?
    \answerNA{}
\end{enumerate}

\item If you ran experiments...
\begin{enumerate}
  \item Did you include the code, data, and instructions needed to reproduce the main experimental results (either in the supplemental material or as a URL)?
    \answerYes{}
  \item Did you specify all the training details (e.g., data splits, hyperparameters, how they were chosen)?
    \answerYes{}
        \item Did you report error bars (e.g., with respect to the random seed after running experiments multiple times)?
    \answerNo{}
    
    Experiments are conducted based on the code released by CEC (CVPR'21), which specified the random seed for re-implementation and comparison. Therefore, we follow the same random seed to conduct experiments and compare with other methods without the error bar.
    
        \item Did you include the total amount of compute and the type of resources used (e.g., type of GPUs, internal cluster, or cloud provider)?
    \answerYes{}
\end{enumerate}

\item If you are using existing assets (e.g., code, data, models) or curating/releasing new assets...
\begin{enumerate}
  \item If your work uses existing assets, did you cite the creators?
    \answerYes{}
  \item Did you mention the license of the assets?
    \answerYes{}
  \item Did you include any new assets either in the supplemental material or as a URL?
    \answerNA{}
  \item Did you discuss whether and how consent was obtained from people whose data you're using/curating?
    \answerYes{}
  \item Did you discuss whether the data you are using/curating contains personally identifiable information or offensive content?
    \answerYes{}
\end{enumerate}

\item If you used crowdsourcing or conducted research with human subjects...
\begin{enumerate}
  \item Did you include the full text of instructions given to participants and screenshots, if applicable?
    \answerNA{}
  \item Did you describe any potential participant risks, with links to Institutional Review Board (IRB) approvals, if applicable?
    \answerNA{}
  \item Did you include the estimated hourly wage paid to participants and the total amount spent on participant compensation?
    \answerNA{}
\end{enumerate}

\end{enumerate}


\appendix

\section{Appendix for Related Work}

\begin{figure}[t]
	\centering
	\includegraphics[width=0.8\linewidth]{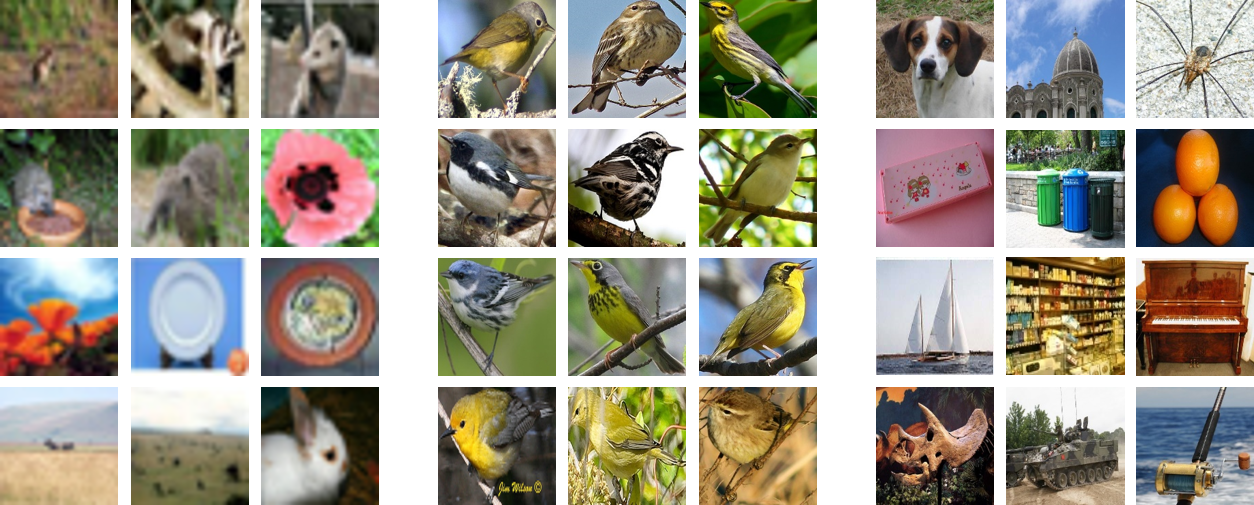}
	\caption{Samples of CIFAR100 (left), CUB200 (mid) and \textit{mini}ImageNet (right).}
	\label{fig:samples}
\end{figure}

\paragraph{Few-Shot Class-Incremental Learning.}

Few-shot class-incremental learning (FSCIL) can be roughly divided into finetune-based methods~\cite{hou2019learning,rebuffi2017icarl,castro2018end,tao2020few} and metric-based methods~\cite{zhang2021few,zhou2022forward}. The former finetunes the model on the novel-class training data, focusing on avoiding catastrophic forgetting during the finetuning. The latter achieves this goal by freezing parameters of the pre-trained model, and recognize the novel classes by the prototype-based~\cite{snell2017prototypical} Nearest-Neighbor classification,
which shares the same concept with the metric-based few-shot learning works~\cite{Vinyals2016Matching,snell2017prototypical,yang2018learning}. 
This paper can be categorized into the metric-based FSCIL methods.

\paragraph{Few-Shot Learning.}

Few-shot Learning (FSL)~\cite{Vinyals2016Matching,snell2017prototypical} focuses on the recognition of novel classes with only few training samples. It can be roughly divided into metric-based methods~\cite{Vinyals2016Matching,snell2017prototypical,zhang2020deepemd,qi2018low}, meta-learning based methods~\cite{finn2017model,nichol2018reptile,rusu2019meta} and augmentation-based methods~\cite{wang2018low,hariharan2017low,bao2020bowtie}. Metric-based methods share the concept with those of FSCIL, which aims to learn a good embedding space to recognize novel classes. Therefore, methods~\cite{qi2018low,liu2020negative,zhang2020deepemd} effective in FSL can also be effective in FSCIL. However, as the original FSL task (except for some subspecies task of FSL, e.g., generalized FSL~\cite{gidaris2018dynamic}) does not require the recognition of base classes, FSL generally emphasizes more on the novel-class generalization.

\paragraph{Margin-Based Classification.}

Margin-based classification~\cite{sun2020circle,wang2018cosface,deng2019arcface,liu2020negative} has been widely utilized in metric learning. 
For example, \cite{wang2018cosface} and \cite{deng2019arcface} proposed to add a positive margin to the classification to learn a better fine-grained embedding space for face recognition. \cite{liu2020negative} proposed to add a negative margin to benefit the novel-class recognition. 
However, the reason why the model behaves differently on base classes and novel classes has not been fully studied, and this paper tackles this problem from the aspect of pattern learning.

\paragraph{Discriminability vs. Transferability.}

The dilemma between transferability and discriminability has been researched in various research domains~\cite{chen2019transferability,cui2022discriminability,cui2020towards}. 
For example, \cite{chen2019transferability} studied this problem for adversarial domain adaptation, \cite{cui2022discriminability} studied this problem from the aspect of information-bottleneck theory, and \cite{cui2020towards} investigated this problem under the label insufficient situations. 
However, it still remains to be studied under the metric-based FSCIL or FSL scenario, and this paper study this problem from the aspect of pattern learning.

\paragraph{Semantic Pattern Learning.}

Interpretability of deep networks~\cite{zhou2016learning,bau2017network} shows that each channel in the extracted feature can be understood as a pattern extractor, which can be used to dissect~\cite{bau2017network} the given network for the encoded knowledge. Moreover, each convolution kernel, as a pattern extractor, can be understood as a semantic template~\cite{chen2020addernet}, and the activation on each channel can be viewed as the matching score between the template and the input. Based on these studies, \cite{zou2020compositional} proposed to view each class as a composition of semantic patterns. Based on the above previous works, we can analyze the model behavior from the aspect of pattern learning.

\begin{table}[t]
	\caption{Comparison with state-of-the-art works on the CUB200 dataset.}
	\label{tab:cub_sota}
	\centering
	\resizebox{1.0\textwidth}{!}{\begin{tabular}{lccccccccccc}
			\toprule
			Method & S0 & S1 & S2 & S3 & S4 & S5 & S6 & S7 & S8 & S9 & S10 \\
			\midrule
			Finetune & 68.68 & 43.70 & 25.05 & 17.72 & 18.08 & 16.95 & 15.10 & 10.06 & 8.93 & 8.93 & 8.47 \\
			Rebalancing~\cite{hou2019learning} & 68.68 & 57.12 & 44.21 & 28.78 & 26.71 & 25.66 & 24.62 & 21.52 & 20.12 & 20.06 & 19.87 \\
			iCaRL~\cite{rebuffi2017icarl} & 68.68 & 52.65 & 48.61 & 44.16 & 36.62 & 29.52 & 27.83 & 26.26 & 24.01 & 23.89 & 21.16 \\
			EEIL~\cite{castro2018end} & 68.68 & 53.63 & 47.91 & 44.20 & 36.30 & 27.46 & 25.93 & 24.70 & 23.95 & 24.13 & 22.11 \\
			TOPIC~\cite{tao2020few} & 68.68 & 62.49 & 54.81 & 49.99 & 45.25 & 41.40 & 38.35 & 35.36 & 32.22 & 28.31 & 26.26 \\
			Decoupled-NegCosine~\cite{liu2020negative} & 74.96 & 70.57 & 66.62 & 61.32 & 60.09 & 56.06 & 55.03 & 52.78 & 51.50 & 50.08 & 48.47 \\
			CEC~\cite{zhang2021few} & 75.85 & 71.94 & 68.50 & 63.50 & 62.43 & 58.27 & 57.73 & 55.81 & 54.83 & 53.52 & 52.28 \\
			FSLL+SS~\cite{mazumder2021few} & 75.63 & 71.81 & 68.16 & 64.32 & 62.61 & 60.10 & 58.82 & 58.70 & 56.45 & 56.41 & 55.82 \\
			FACT~\cite{zhou2022forward} & 75.90 & 73.23 & 70.84 & 66.13 & 65.56 & 62.15 & 61.74 & 59.83 & 58.41 & 57.89 & 56.94 \\
			IDLVQ-C~\cite{chen2020incremental} & 77.37 & 74.72 & 70.28 & 67.13 & 65.34 & 63.52 & 62.10 & 61.54 & 59.04 & 58.68 & 57.81 \\
			\midrule
			Ours & \textbf{79.57} & \textbf{76.07} & \textbf{72.94} & \textbf{69.82} & \textbf{67.80} & \textbf{65.56} & \textbf{63.94} & \textbf{62.59} & \textbf{60.62} & \textbf{60.34} & \textbf{59.58} \\
			\bottomrule
	\end{tabular}}\vspace{-0.4cm}
\end{table}

\begin{table}[t]
	\caption{Comparison of state-of-the-art works on the CIFAR100 dataset.}
	\label{tab:cifar_sota}
	\centering
	\resizebox{0.88\textwidth}{!}{\begin{tabular}{lccccccccc}
			\toprule
			Method & S0 & S1 & S2 & S3 & S4 & S5 & S6 & S7 & S8 \\
			\midrule
			Finetune & 64.10 & 39.61 & 15.37 & 9.80 & 6.67 & 3.80 & 3.70 & 3.14 & 2.65 \\
			Pre-Allocated RPC~\cite{pernici2021class} & 64.50 & 54.93 & 45.54 & 30.45 & 17.35 & 14.31 & 10.58 & 8.17 & 5.14 \\
			iCaRL~\cite{rebuffi2017icarl} & 64.10 & 53.28 & 41.69 & 34.13 & 27.93 & 25.06 & 20.41 & 15.48 & 13.73 \\
			EEIL~\cite{castro2018end} & 64.10 & 53.11 & 43.71 & 35.15 & 28.96 & 24.98 & 21.01 & 17.26 & 15.85 \\
			Rebalancing~\cite{hou2019learning} & 64.10 & 53.05 & 43.96 & 36.97 & 31.61 & 26.73 & 21.23 & 16.78 & 13.54 \\
			TOPIC~\cite{tao2020few} & 64.10 & 55.88 & 47.07 & 45.16 & 40.11 & 36.38 & 33.96 & 31.55 & 29.37 \\
			Decoupled-NegCosine~\cite{liu2020negative} & 74.36 & 68.23 & 62.84 & 59.24 & 55.32 & 52.88 & 50.86 & 48.98 & 46.66 \\
			Decoupled-Cosine~\cite{Vinyals2016Matching} & \textbf{74.55} & 67.43 & 63.63 & 59.55 & 56.11 & 53.80 & 51.68 & 49.67 & 47.68 \\
			Decoupled-DeepEMD~\cite{zhang2020deepemd} & 69.75 & 65.06 & 61.20 & 57.21 & 53.88 & 51.40 & 48.80 & 46.84 & 44.41 \\
			CEC~\cite{zhang2021few} & 73.07 & 68.88 & 65.26 & 61.19 & 58.09 & 55.57 & 53.22 & 51.34 & 49.14 \\
			\midrule
			CLOM (Ours) & 74.20 & \textbf{69.83} & \textbf{66.17} & \textbf{62.39} & \textbf{59.26} & \textbf{56.48} & \textbf{54.36} & \textbf{52.16} & \textbf{50.25} \\
			\bottomrule
	\end{tabular}}
\end{table}

\begin{table}[t]
	\caption{Comparison of state-of-the-art works on the \textit{mini}ImageNet dataset.}
	\label{tab:mini_sota}
	\centering
	\resizebox{0.88\textwidth}{!}{\begin{tabular}{lccccccccc}
			\toprule
			Method & S0 & S1 & S2 & S3 & S4 & S5 & S6 & S7 & S8 \\
			\midrule
			Finetune & 61.31 & 27.22 & 16.37 & 6.08 & 2.54 & 1.56 & 1.93 & 2.60 & 1.40 \\
			Pre-Allocated RPC~\cite{pernici2021class} & 61.25 & 31.93 & 18.92 & 13.90 & 14.37 & 15.57 & 16.15 & 12.33 & 12.28 \\
			iCaRL~\cite{rebuffi2017icarl} & 61.31 & 46.32 & 42.94 & 37.63 & 30.49 & 24.00 & 20.89 & 18.80 & 17.21 \\
			EEIL~\cite{castro2018end} & 61.31 & 46.58 & 44.00 & 37.29 & 33.14 & 27.12 & 24.10 & 21.57 & 19.58 \\
			Rebalancing~\cite{hou2019learning} & 61.31 & 47.80 & 39.31 & 31.91 & 25.68 & 21.35 & 18.67 & 17.24 & 14.17 \\
			TOPIC~\cite{tao2020few} & 61.31 & 50.09 & 45.17 & 41.16 & 37.48 & 35.52 & 32.19 & 29.46 & 24.42 \\
			Decoupled-NegCosine~\cite{liu2020negative} & 71.68 & 66.64 & 62.57 & 58.82 & 55.91 & 52.88 & 49.41 & 47.50 & 45.81 \\
			Decoupled-Cosine~\cite{Vinyals2016Matching} & 70.37 & 65.45 & 61.41 & 58.00 & 54.81 & 51.89 & 49.10 & 47.27 & 45.63 \\
			Decoupled-DeepEMD~\cite{zhang2020deepemd} & 69.77 & 64.59 & 60.21 & 56.63 & 53.16 & 50.13 & 47.79 & 45.42 & 43.41 \\
			CEC~\cite{zhang2021few} & 72.00 & 66.83 & 62.97 & 59.43 & 56.70 & 53.73 & 51.19 & 49.24 & 47.63 \\
			\midrule
			CLOM (Ours) & \textbf{73.08} & \textbf{68.09} & \textbf{64.16} & \textbf{60.41} & \textbf{57.41} & \textbf{54.29} & \textbf{51.54} & \textbf{49.37} & \textbf{48.00} \\	
			\bottomrule
	\end{tabular}}
\end{table}

\section{Appendix for Experiments}

\subsection{Detailed Dataset Description}

\paragraph{CIFAR100.} CIFAR100~\cite{krizhevsky2009learning} is a challenging dataset consisting of 100 classes and 60,000 images with the shape of 32 $\times$ 32 as shown in Fig.~\ref{fig:samples} (left). We adopt this dataset with consent from the authors~\cite{krizhevsky2009learning}. 
For each class, there are 500 images for training and 100 images for testing. As split in \cite{tao2020few}, 60 classes are chosen as the base-session classes, and 40 classes are used as novel classes. The 40 novel classes are further divided into 8 incremental sessions where each session has 5 classes with 5 training samples in each class for training.

\paragraph{Caltech-UCSD Birds-200-2011 (CUB200).} The CUB200~\cite{wah2011caltech} dataset is designed for the fine-grained classification of birds as shown in Fig.~\ref{fig:samples} (mid). 
We adopt this dataset with consent from the authors~\cite{wah2011caltech}.
It contains 11,788 images from 200 classes. As split in \cite{tao2020few}, 100 classes are chosen as the base-session classes and the remaining are the novel classes. The 100 novel classes are further divided into 10 incremental sessions where each session contains 10 classes with 5 training samples in each class.

\paragraph{\textit{mini}ImageNet.} The \textit{mini}ImageNet~\cite{Vinyals2016Matching} dataset is a subset of the ImageNet~\cite{deng2009imagenet} dataset, containing 100 classes and 600 images in each class, and the images of it are resized to 84 $\times$ 84 as shown in Fig.~\ref{fig:samples} (right).
We adopt this dataset with consent from the authors~\cite{Vinyals2016Matching}.
We follow \cite{tao2020few} to split it into 60 base classes and 40 novel classes, and construct 8 incremental sessions from the 40 novel classes, where each session contains 5 classes with 5 training samples in each class.

\subsection{Detailed Implementation Details}

Our implementation is based on the code released by CEC~\cite{zhang2021few} under the MIT license.

\paragraph{CIFAR100.} We follow CEC~\cite{zhang2021few} to utilize ResNet20~\cite{he2016deep} as the backbone network. The data augmentation includes regular augmentation techniques, i.e., the random resized crop, the random horizontal flip and the normalization of images. Note that for fair comparison, we do not utilize the auto-augment tricks as done in \cite{zhou2022forward}. We follow CEC to train the model for 100 epochs, and decay the original learning rate 0.1 to 0.01 and 0.001 at the 60th and 70th epoch respectively. Other details have been illustrated in the paper.

\paragraph{CUB200.} ResNet18~\cite{he2016deep} is utilized as the backbone network following CEC. Also, the pre-training from ImageNet is adopted as CEC and TOPIC~\cite{tao2020few}. Therefore, we shrink the learning rate of the backbone network to 0.01, while keeping the that for the global learning rate to be 0.1. The model is trained for 80 epochs, and the learning rate decay is conducted at the 40th and 50th epoch. Data augmentation techniques are the same as those in CIFAR100. Other details have been illustrated in the paper.

\paragraph{\textit{mini}ImageNet.} ResNet18 is also utilized as the backbone network following CEC. Unlike that in CUB200, the backbone is trained from scratch. The data augmentation is also the same as that on CIFAR100, where the auto-augment is not adopted neither. The model is trained for 180 epochs, and the learning rate is decayed to 10\% at the 90th and the 120th epoch. Other details have been illustrated in the paper.

\subsection{Detailed Incremental Performance}

In the paper, we plot the comparison on CUB200 and \textit{mini}ImageNet by the performance curve. For clarity, we also report the exact numbers in Tab.~\ref{tab:cifar_sota} and \ref{tab:mini_sota}, and attach the results on CUB200 in Tab.~\ref{tab:cub_sota} for easy comparison.

From these tables, we can see that our performance measured by the last incremental session is the highest, even with lower base-class performance as in Tab.~\ref{tab:cifar_sota}. This result verifies that our model could achieve better novel-class generalization without harming the base-class performance, i.e., mitigating the class-level overfitting problem.

\subsection{Extra Ablation Study}

\paragraph{Pattern Cooperation}

\begin{figure}[t]
	\centering
	\includegraphics[width=0.4\linewidth]{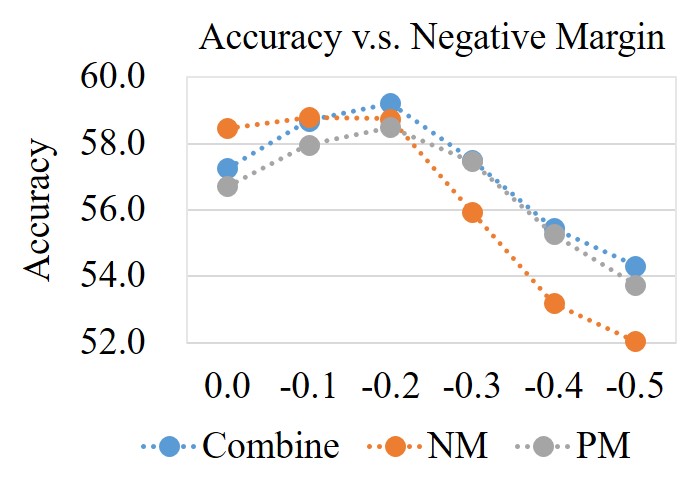}
	\centering
	\includegraphics[width=0.4\linewidth]{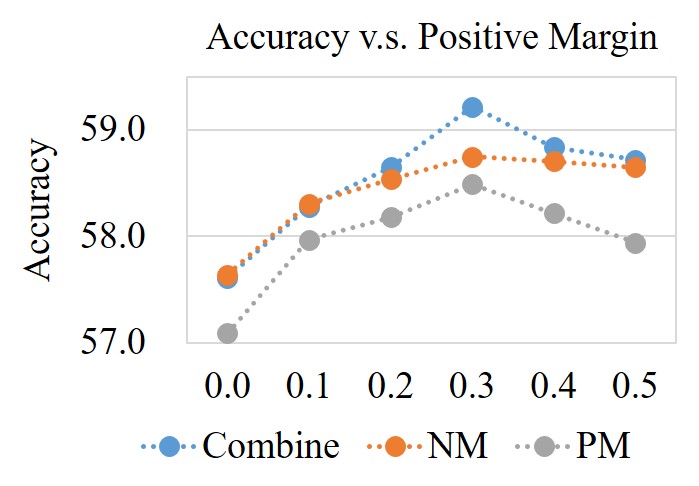}
	\caption{Performance of the NM feature, the PM feature and the combined feature w.r.t. the negative (left) and positive (right).}
	\label{fig:cooperation}
\end{figure}

To verify the learning of positive- and negative-margin-based features, we study the performance on different features w.r.t. margins without the relation mapping module. Results are reported in Fig.~\ref{fig:cooperation} (left) and Fig.~\ref{fig:cooperation} (right), where \textit{NM} denotes the negative-margin-based feature, \textit{PM} positive-margin-based feature, and \textit{Combine} refers to the concatenation of two features. The experiments are conducted with the margin of the other branch fixed. We can see that 


\textbf{(1) Branches learn to encode different information as the difference of margin between two features increases}. As can be seen, when the margin is 0.0, the performance of the combined feature is lower than one of the branches. As the margin decreases (increases) in the left (right) figure, the performance of the combined feature begin to surpass both two branches, which validates that two branches learn to encode different information (i.e., one for transferability and one for discriminability) as the difference of margins increases.  

\textbf{(2) The improvements on one branch can also help the learning of the other branch}, as the performance of two branches increases and decreases synchronously. This is because the PM feature is built from the NM feature, therefore the learning of one branch can implicitly help the other one.

\paragraph{Sensitivity Study}

\begin{figure}[t]
	\centering
	\includegraphics[width=0.4\linewidth]{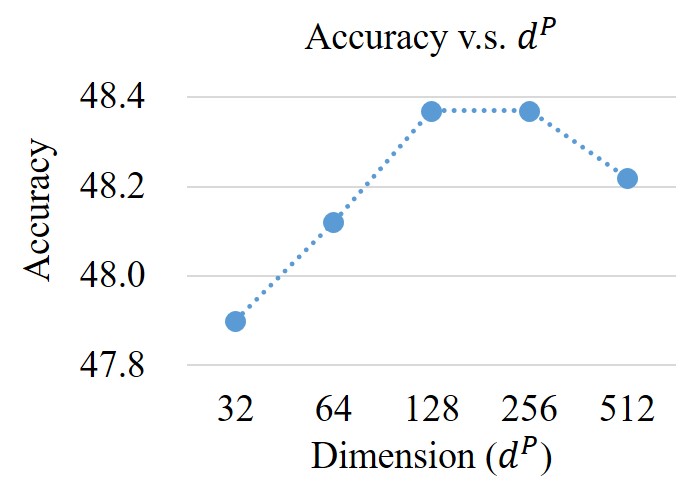}
	\centering
	\includegraphics[width=0.4\linewidth]{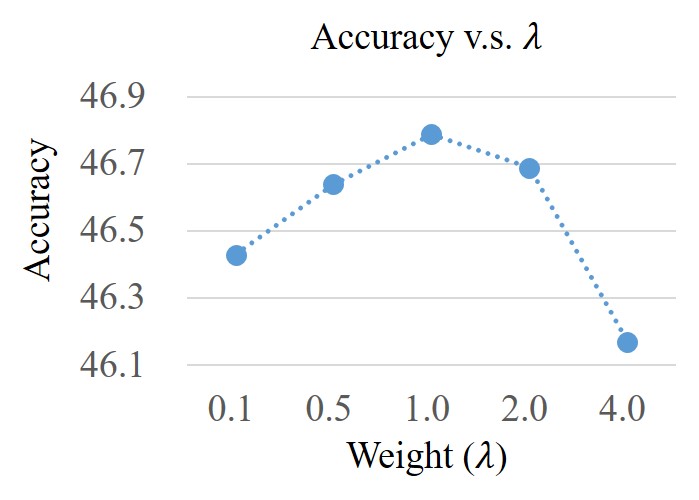}
	\caption{Left: sensitivity study of the PM feature dimension ($d^P$). Right: sensitivity study of the weight for PM feature classification ($\lambda$).}
	\label{fig:sensitivity}
\end{figure}

To verify the choice of hyper-parameters, we plot the sensitivity study of the PM feature's dimension $d^P$, and the weight $\lambda$ of the classification loss for the PM feature. Results are plotted in Fig.~\ref{fig:sensitivity}. We can see that $d^P$ reaches its maximum performance on CIFAR100 at the chosen optimal value, i.e., 256, and drops when keeping increasing it. Other datasets follow the same trend for the chosen values, i.e., 8192 for CUB200 and 4096 for \textit{mini}ImageNet. 

Similarly, the overall accuracy reaches the top value when $\lambda$ equals 1.0, which means the contribution from NM and PM features are close. Other datasets also follow the same trend for the chosen values. Moreover, note that the best $\lambda$ for CUB200 is 0.01, this is because the backbone model adopts the pre-training~\cite{tao2020few,zhang2021few} from ImageNet, therefore the guidance from PM patterns on NM patterns should be weakened.

\section{Full combinations of different margins}

We report the experiments with all possible combinations of margins on CIFAR100 in Tab.~\ref{tab:margin_combinations}, where the horizontal axis represents the margin attached to higher layer $F(\cdot)$, and the vertical axis denotes the margin for the lower layer $f(\cdot)$.

\begin{table}[t]
	\caption{Full combinations of different margins without the relation mapping module (CIFAR100).}
	\label{tab:margin_combinations}
	\centering
	\resizebox{0.88\textwidth}{!}{\begin{tabular}{c|cccccccccc}
			\toprule
			Margin & -0.5 &	-0.4 &	-0.3 &	-0.2 &	-0.1 &	0.0 &	0.1 & 	0.2  &	0.3  &	0.4 \\
			\midrule
			-0.5 &	41.16 &	42.51 &	43.30 & 45.68 &	47.93 &	49.33 &	49.25 &	49.27 &	48.53 &  48.01 \\
			-0.4 &	42.24 &	42.37 &	44.37 &	45.96 &	48.17 &	48.93 &	48.94 &	48.76 &	49.01 &  47.15 \\
			-0.3 &	43.99 &	44.45 &	45.06 &	45.85 &	47.51 &	48.90 &	49.38 &	49.09 &	48.87 &	 48.08 \\
			-0.2 &	45.05 &	44.71 &	45.61 &	45.85 &	48.20 &	48.08 &	49.60 &	48.65 &	48.53 &  48.08 \\
			-0.1 &	47.50 &	47.11 &	46.71 &	46.69 &	47.99 &	48.87 &	48.92 &	49.21 &	48.55 &  47.99 \\
			0.0	 &  47.38 &	47.7  &	47.00 &	47.94 &	47.43 &	48.48 &	48.67 &	49.14 &	47.96 &  47.87 \\
			0.1  &	47.32 &	48.14 &	48.01 &	47.96 &	47.95 &	48.37 &	47.62 &	48.02 &	48.35 &  47.32 \\
			0.2  &	47.44 &	46.66 &	47.7  &	47.05 &	47.27 &	47.21 &	47.40 &	47.55 &	47.64 &  46.65 \\
			0.3  &	46.61 &	46.23 &	46.29 &	46.21 &	47.01 &	47.11 &	46.73 &	46.37 &	46.55 &  45.68 \\ 
			0.4  &	45.37 &	45.35 &	45.81 &	45.70 &	46.26 &	46.69 &	45.79 &	46.51 &	46.08 &  45.56 \\
			\bottomrule
	\end{tabular}}
\end{table}

We can see that the margins adopted in the paper (i.e., positive margin on the higher layer + negative margin on the lower layer, the top right area) show clear improvements over the baseline (i.e., margins on both layers are 0.0). Instead, other combinations of margins (e.g., negative margin on the higher layer + positive margin on the lower layer, the left bottom area) show a much lower performance compared with the baseline, which verifies our choice of hyper-parameters and our insight: build positive-margin-based patterns from negative-margin-based patterns.

\section{Comparison with face recognition methods}

We implemented some methods in the face recognition community that might be relevant to our work, including

(1) Adaptive-margin-based methods (CurricularFace~\cite{huang2020curricularface}, AdaCos~\cite{zhang2019adacos}, ElasticFace~\cite{boutros2022elasticface}, AMR-Loss~\cite{zhang2021face}), such as adjusting the margin value according to the intra/inter-class angles. However, these methods always rely more on the angles across the training time than on angles across all training classes, which differs from our class-relation-based mapping mechanism which rely more on angles across all training classes.

(2) Relational-margin-based methods (TRAML~\cite{li2020boosting}), which maps the class relationship to the margin. However, this work takes the semantic embedding (such as attributes) of each class as input, and utilizes a network to learn the relational margin, which differs from our work in that we do not need further training a network nor the attributes to obtain the relational margin. Moreover, we specifically design the search space of hyper-parameters of the linear mapping to allow similar classes to have a margin with larger absolute value, so that the mapping is more interpretable than mapping by a learnable black-box neural network.

Below we empirically compare our method with the above face recognition methods on CIFAR100 following settings in our paper. Our aim of experiments includes (1) verifying whether they can solve the dilemma between base-class performance and novel-class generalization and (2) verifying whether they can effectively capture the relationship between classes.

\begin{table}[h]
	\caption{Comparison with face recognition methods on CIFAR100.}
	\label{tab:face_cifar}
	\centering
	\resizebox{0.7\textwidth}{!}{\begin{tabular}{lccc}
			\toprule
			CIFAR100	                 		&	overall &	novel 	&	base	\\
			\midrule
			baseline                  			&	47.02 	&	37.40  	&	72.32	\\
			baseline + CurricularFace    		&	47.22 	&	35.77 	&	72.67	\\
			baseline + AdaCos	          		&	44.48 	&	26.07 	&	72.55	\\
			baseline + TRAML 	          		&	47.31 	&	36.32 	&	73.15	\\
			baseline + ElasticFace  	   		&	47.09 	&	36.60  	&	72.40	\\
			baseline + AMR-Loss          		&	46.66 	&	33.37 	&	72.58	\\
			baseline + NM/PM (ours)     		&	\textbf{49.21} 	&	\textbf{40.22} 	&	\textbf{73.72}	\\
			baseline + NM/PM + CurricularFace 	&	49.48 	&	40.07 	&	74.10	\\
			baseline + NM/PM + AdaCos     		&	45.24 	&	27.20  	&	73.83	\\
			baseline + NM/PM + TRAML      		&	48.18 	&	39.35 	&	73.73	\\
			baseline + NM/PM + ElasticFace   	&	49.43 	&	40.70  	&	74.09	\\
			baseline + NM/PM + AMR-Loss 		&	49.22 	&	38.92 	&	74.15	\\
			baseline + NM/PM + relation (ours)  &	\textbf{50.25} 	&	\textbf{41.17} 	&	\textbf{74.20}	\\
			\bottomrule
	\end{tabular}}
\end{table}

From this table, we can see that (1) these methods cannot solve the dilemma by adding directly to the baseline method, since none of them can achieve higher base and novel accuracy simultaneously compared with the baseline performance, while ours (baseline + NM/PM) can; (2) with the NM/PM architecture design, these methods can hardly capture the relationship between classes, since they could not achieve performance significantly higher than the baseline + NM/PM ones, which verify the effectiveness of our method under the few-shot class-incremental learning task.

\section{Broader Impact}

We propose a margin-based FSCIL method to mitigate the class-level overfitting problem in the margin-based classification. This work can also be adopted in fields other than FSCIL, such as FSL, image retrieval~\cite{deng2019arcface} and person re-identification~\cite{wang2021batch}, because the class-level overfitting (CO) problem handled by this method is not limited to the FSCIL task. 
The limitation of the work is to omit the many-shot scenarios where the finetuning on novel classes cannot be ignored. However, as the novel-class embedding extracted by our method could provide an effective initialization for the novel-class classifier, this method still has the potential to be further developed for the realistic many-shot scenarios.

\end{document}